\documentclass{article} % For LaTeX2e
\usepackage{iclr2027_conference,times}
\let\cite\citep
\usepackage{amsmath,amsfonts,bm}

\def\eqref#1{equation~\ref{#1}}
\def\1{\bm{1}}

\DeclareMathAlphabet{\mathsfit}{\encodingdefault}{\sfdefault}{m}{sl}
\SetMathAlphabet{\mathsfit}{bold}{\encodingdefault}{\sfdefault}{bx}{n}

\usepackage{url}
\usepackage{graphicx}
\usepackage[most]{tcolorbox}
\usepackage{amsmath}
\usepackage{placeins}
\usepackage{capt-of}
\usepackage[utf8]{inputenc} % allow utf-8 input
\usepackage[T1]{fontenc}    % use 8-bit T1 fonts

\usepackage{url}            % simple URL typesetting
\usepackage{booktabs}       % professional-quality tables
\usepackage{longtable}      % multi-page tables
\usepackage{amsfonts}       % blackboard math symbols
\usepackage{nicefrac}       % compact symbols for 1/2, etc.
\usepackage{microtype}      % microtypography
\usepackage{xcolor}         % colors
\usepackage{hyperref}
\usepackage{float}
\usepackage[capitalize]{cleveref}
\crefname{appendix}{Appendix}{Appendices}
\Crefname{appendix}{Appendix}{Appendices}

\usepackage{array}
\newcolumntype{L}[1]{>{\raggedright\arraybackslash}p{#1}}

\title{One Sentence, One Drama: Personalized Short-Form Drama Generation via Multi-Agent Systems}

\author{
Yufei Shi$^{1}$\thanks{Equal contribution.} \quad
Weilong Yan$^{2*}$ ~\quad
Naixuan Huang$^{1}$ ~\quad
Yucheng Chen$^{1}$ ~\quad
Chenyu Zhang$^{3}$ \\
\textbf{Yiming Cheng}$^{4}$ ~\quad
\textbf{Tao He}$^{5}$ ~\quad
\textbf{Si Yong Yeo}$^{1}$\thanks{Corresponding author.}  ~\quad
\textbf{Ming Li}$^{6\dagger}$ 
\\[3pt]
$^{1}$MedVisAI Lab, Lee Kong Chian School of Medicine, Nanyang Technological University \quad \\
$^{2}$NUS \quad
$^{3}$BIT \quad
$^{4}$THU \quad
$^{5}$UESTC \quad
$^{6}$CUHKSZ
\vspace{-3mm}
}

\iclrfinalcopy % Uncomment for camera-ready version, but NOT for submission.
\begin{document}

\maketitle

\begin{abstract}
Existing approaches for digital short-drama production typically rely on one-shot LLM generated scripts and loosely coupled pipelines, which fail to satisfy three key requirements of short-drama generation: (1) narrative pacing, resulting in weak hooks, insufficient escalation, and unattractive endings; (2) spatial consistency, leading to drifting scene layouts and inconsistent character positions across clips; and 
(3) production-level quality control, requiring extensive manual review and correction across script and visual stages.
We present \textit{One Sentence, One Drama}, a hierarchical multi-agent framework that transforms a user's single-sentence idea into a fully produced short drama through structured intermediate modules and iterative refinement. Our approach is built upon three key components: (1) a multi-agent debate-based story generation module that enforces short-drama pacing and narrative coherence; (2) a 3D-grounded first-frame generation mechanism that establishes a shared spatial reference for consistent character positioning and scene layout across clips; and (3) multi-stage reviewer loops that perform comprehensive error detection and targeted revision across script, visual, and video generation stages. We also introduce scene-level BGM matching and scene transition planning to improve the audience's immersive experience.
To systematically evaluate this task, we introduce \textit{Short-Drama-Bench}, a benchmark that extends standard video quality metrics with short-drama-specific criteria. Experimental results demonstrate that our method significantly outperforms existing pipelines in narrative quality, cross-clip consistency, and overall viewing experience.
\end{abstract}

\section{Introduction}
\vspace{-2mm}
Recent advances in video foundation models have substantially improved automated short-clip generation. Models such as Sora~\cite{sora}, Seedance~\cite{seedance2.0}, Kling~\cite{kling}, and Veo~\cite{veo3} have demonstrated strong capabilities in visual fidelity, motion realism, and prompt following. These models provide a powerful basis for generating high-quality video clips from textual or visual conditions. Recent long-form generation pipelines have explored combining large language model planning with video synthesis. Systems such as MovieAgent~\cite{movieagent}, StoryMem~\cite{storymem}, and ScriptAgent~\cite{scriptagent} decompose long-video creation into multiple stages, representing an important step toward automated long-form video production. Nevertheless, these methods are primarily designed for organizing clips into longer videos and do not explicitly model the distinctive narrative dynamics of short dramas, which demand dense dramatic hooks—characterized by rapid conflict onset, high-frequency escalation and reversals, and fast-paced payoff within a highly compressed duration \cite{cao2026audience}.

More recently, Toonflow~\cite{toonflow} and Xiaoyunque~\cite{xiaoyunque} have adapted generative models to short-drama production workflows. However, they still face three major limitations. First, they often rely on a ready-made story input, which shifts the burden of short-drama writing to the user \cite{cao2026audience}. When only a brief idea is provided, they simply use one-shot LLM expansion, leading to weak dramatic hooks and unsatisfactory storylines. Second, they usually create clips using loosely connected generation units \cite{longcontexttuning,elmoghany2026infinitystory}, causing cross-clip spatial inconsistencies such as drifting scene layouts, abrupt character position changes, and unresolved prop states. Third, their outputs typically require substantial manual inspection and correction across script, keyframe, and video stages before reaching production-level quality, due to diverse errors in pacing, character consistency, dialogue accuracy, spatial layout, prop states, and action continuity \cite{longcontexttuning,elmoghany2026infinitystory,holocine}.

To address these challenges, we present \textit{One Sentence, One Drama}, a hierarchical multi-agent framework for generating an entire short drama from a single-sentence idea. Our framework decomposes the generation process into a hierarchy of structured and revisable intermediate modules.
Specifically, our framework consists of three core components. First, we introduce a \textbf{multi-agent debate-based story generation} module that improves short-drama pacing and narrative coherence by explicitly modeling opening hooks, conflict escalation, ending suspense, and storyline consistency through synergistic debate and revision. To counter the over-correction of iterative revision, which tends to delete memorable hooks and reversals, an \textbf{Idea Bank} records the removed elements and selectively revives them in a final round. Second, we propose \textbf{3D-grounded first-frame generation} to address cross-clip spatial drift. By constructing a scene-level 3D world model and aligning frames within a shared spatial coordinate system, the method enables consistent character positioning and scene layout across clips, even under severe viewpoint changes or scene re-establishment. We design \textbf{geometry-adaptive character registration} to place the single-frame character mesh into the scene-level 3D world at the correct size and distance. Third, we design \textbf{multi-stage reviewer loops} across script, prompt, keyframe, and video generation to enforce constraints on pacing, spatial relations, prop states, physical plausibility, and action continuity. In addition, we incorporate scene-level BGM matching and transition planning to further enhance the immersive viewing experience.

To verify our framework, we introduce \textit{Short-Drama-Bench}, a novel and challenging benchmark that augments standard video-quality metrics with short-drama-specific criteria, including narrative engagement, spatial continuity, and full-production viewing experience. 
It consists of $50$ diverse story prompts spanning $7$ popular categories (e.g., Time Travel \& Rebirth, Suspense \& Thriller, Sweet Romance) and $17$ fine-grained subcategories. Each subcategory contains $2$–$3$ representative samples, covering a broad range of commonly observed short-drama patterns and narrative structures. 

To further reflect the practical complexity of this task, we generate full short-drama outputs for all benchmark prompts, resulting in a total of approximately $239$ minutes of video content. The generated results include a mixture of long-, medium-, and short-duration dramas, consisting of $2$ long-form dramas ($\approx 30$ minutes each), $5$ medium-length dramas ($\approx 10$ minutes each), and $43$ short dramas ($\approx 3$ minutes each). This large-scale generation setup highlights the long-horizon consistency challenges of the task, as models must maintain narrative coherence, character consistency, and spatial continuity across hundreds of sequential clips. These characteristics make Short-Drama-Bench significantly more demanding than conventional short video benchmarks that focus on isolated clip generation.
Experimental results demonstrate that our agentic framework consistently outperforms existing generation pipelines in narrative quality, cross-clip consistency, and overall viewing experience.

In summary, our main contributions in this work are as follows:
\begin{itemize}

\item We formulate single-sentence short-drama generation as a structured generation problem that requires jointly modeling narrative pacing, spatial consistency, and production-level coherence. We propose \textit{One Sentence, One Drama}, a hierarchical multi-agent framework that transforms one-shot generation into a controllable and self-refining process.
\vspace{-1mm}

\item We introduce two key technical innovations to address the core challenges of this task: (i) a multi-agent debate-based story generation module, with an Idea Bank with final-round revival, to improve short-drama pacing and narrative coherence, and (ii) 3D-grounded first-frame generation,  with geometry-adaptive character registration, to enforce cross-clip spatial consistency via a shared spatial coordinate system.
\vspace{-1mm}

\item We present \textit{Short-Drama-Bench}, a diverse and challenging benchmark with $50$ prompts across $7$ categories and $17$ subcategories, along with short-drama-specific evaluation metrics. Our benchmark enables systematic evaluation of narrative quality, spatial continuity, and full-production viewing experience.

\end{itemize}

\section{Personalized Short-Form Drama Generation}
\begin{figure}[t]
\centering
\includegraphics[width=1.0\textwidth]{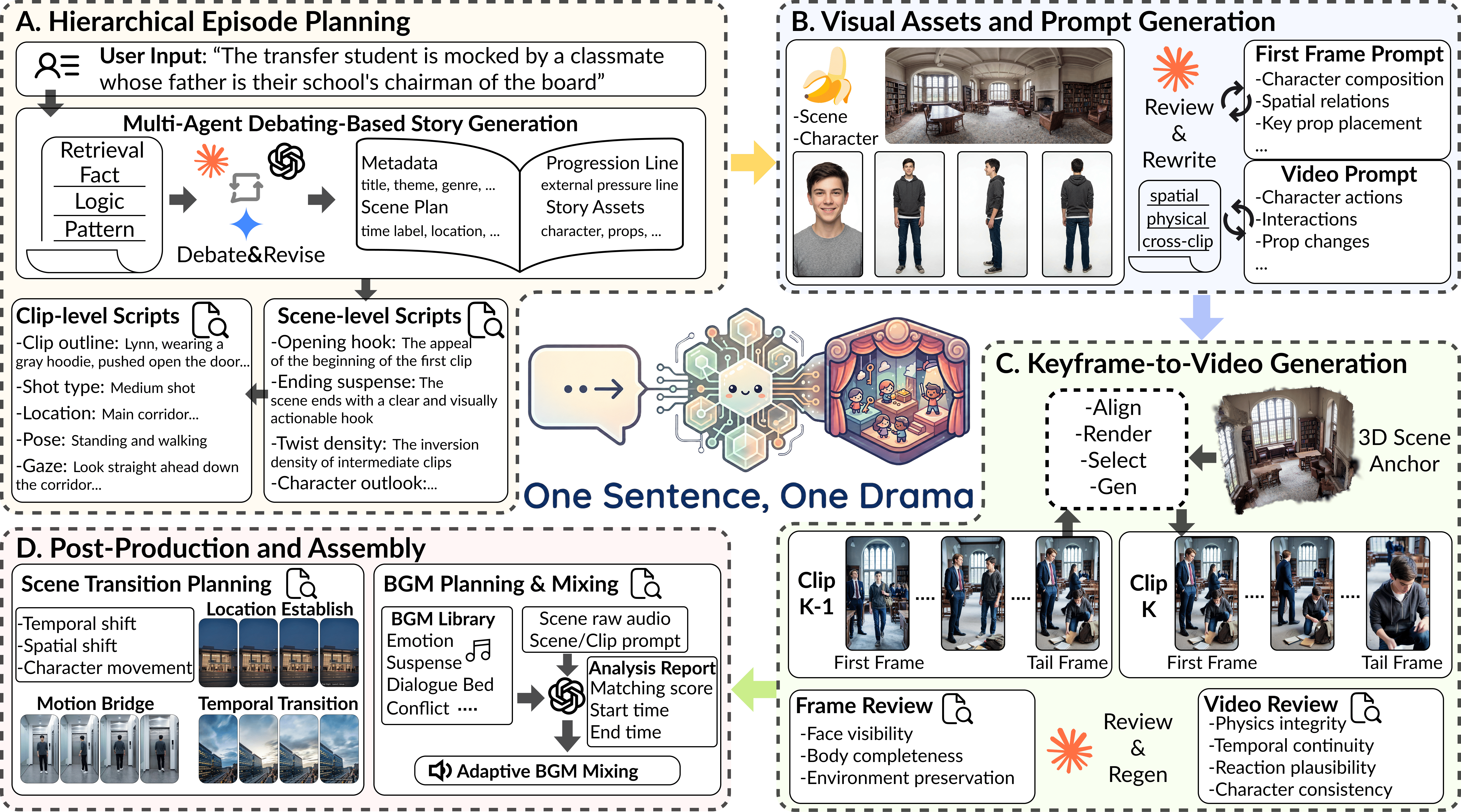}
\vspace{-5mm}
\caption{
Overview of our personalized short-form drama generation with four stages. 
Given a single-sentence input, the system generates structured story and clip scripts through retrieval and multi-agent debate (A), expands into visual assets and paired first-frame/video prompts with prompt review (B), produces 3D-anchored keyframe-to-video clips with frame and video review loops (C), and assembles the final drama with scene transition planning and adaptive BGM mixing (D).
}
\vspace{-5mm}
\label{overview}
\end{figure}

\cref{overview} shows our hierarchical sentence-to-video pipeline. A single-sentence input is transformed into structured story plans and scene/clip-level scripts (\cref{overview}.A), scene-level visual assets and paired prompts (\cref{overview}.B), 3D-anchored keyframe-to-video generation (\cref{overview}.C), and post-production with scene transitions and BGM (\cref{overview}.D). Reviewer loops are inserted across stages for quality control and cross-stage consistency.
\cref{sec:Hierarchical Episode Plan} describes episode planning with atom corpus construction, retrieval, and multi-agent debate-based story generation. \cref{sec:visual assets and prompt generation} presents visual assets and prompt generation. \cref{sec: keyframe to video generation} introduces our strategy for 3D-grounded next-keyframe and next-clip generation. \cref{sec:post production and assembly} illustrates the cross-clip transition planning and drama BGM mixing.

\subsection{Hierarchical Episode Planning}
\label{sec:Hierarchical Episode Plan}

\textbf{Atom Script Corpus Construction.}
Directly expanding a short-drama script from a single prompt often causes weak pacing and unstable local logic.
To address this, we build an atomized corpus from $1{,}102$ high-performing short-drama scripts and construct two retrieval banks. 
Each script is distilled into a structured script card, and together, these cards contain $16{,}311$ beat-level units, encoding cues such as opening action, conflict function, and closing hook visual. 
These embedded beats form the \emph{Pattern Bank}, which provides reusable pacing and dramatic packaging priors. 
In parallel, we split scripts into overlapping local chunks to form the \emph{Logic Bank}, preserving causal context such as motivations, evidence activation, consequence transitions, and scene continuity.
Thus, the corpus is transformed into transferable patterns and logic atoms rather than copied directly. Details are provided in \cref{app:narrative_memory_units}.

\textbf{Multi-Agent Debating-Based Story Generation.}
Given a user's sentence as a logline, we first expand it into a seed text containing a preliminary story skeleton. Based on these, an LLM produces a problem-driven retrieval plan with three routes: fact, logic, and pattern. Fact retrieval invokes web search for externally constrained content, such as law, medicine, and history. Logic retrieval queries the \emph{Logic Bank} for local causal support, while pattern retrieval queries the \emph{Pattern Bank} for relevant short-drama structures. The retrieved references are summarized into fact, logic, and pattern atoms, providing factual, causal, and pacing priors for story drafting.
Combining all these, the pipeline generates a structured story core, containing story-level metadata and the scene plan.

Next, we introduce scene-level script review and rewrite through a multi-agent debating loop. The draft story, story core, and retrieval atoms are reviewed by three independent LLM judges.
When these judges provide conflicting revision suggestions, we send these suggestions to GPT-5.4 Pro as the final decider. The issues are passed to a reviser for local rewriting rather than full regeneration. 

\textbf{Idea Bank with Final-Round Revival.}
Iterative revision has a systematic bias: each round removes risk, deleting memorable elements such as hooks, reversals, and foreshadowing along the way, so the script becomes safer but flatter. To remedy this over-correction, we record every element deleted or weakened during revision, together with its reason, in an Idea Bank. After the revision loop terminates, a final-round revival step restores those elements that can return without harming narrative logic or visual executability.
% Valuable but removed hooks, reversals, or dramatic ideas are stored in an \emph{Idea Bank} and restored in the final round if they do not harm logic or visual executability.  This turns story generation into an agentic review-and-rewrite process. 
More details are given in \cref{app:multi_agent_debate} and \cref{Appendix multi-agent detail}.

\textbf{Scene-level and Clip-level Script Synthesis.}
After obtaining the story core containing the rewrite scene plan, we synthesize clip-level scripts for visual generation. Each scene is then decomposed into temporally ordered clip-level scripts, where each clip specifies its local narrative description, shot type, characters, key props, dialogue or audio cues, actions, interactions, and so on. We also extract each clip's initial and ending states before visual asset generation.

Finally, a clip-level review and rewrite loop is designed for short-drama pacing. The reviewer evaluates the opening hook, ending suspense, and twist density. Based on the evaluation, we perform partitioned rewriting: the opening-hook review revises only the first clip to ensure opening attraction; the ending-suspense review revises only the last clip to ensure a clear and visually actionable hook that motivates continued viewing; and the twist-density review revises only the middle clips to increase reversals, escalations, or information reveals. This strategy strengthens short-drama rhythm while preserving the story structure and core.

\subsection{Visual Assets and Prompt Generation}
\label{sec:visual assets and prompt generation}

\textbf{Scene-level Visual Assets.}
We expand the structured script into scene-level visual assets for subsequent keyframe generation and video rendering. 
Specifically, for each scene, we generate a $360^\circ$ panorama from the scene description, spatial anchors, and the initial character--prop layout. 
This panorama serves as an environment reference for viewpoint selection and for maintaining cross-clip spatial consistency. 
We construct scene-level character assets, 
%
% Based on the scene-level character outlook, we obtain generated or user-uploaded seed portraits for the major characters, and then produce multi-view character references based on the wardrobe descriptions.
%
which are used to produce multi-view character references based on the wardrobe descriptions.
The optional personalization operation from user-provided portraits is detailed in \cref{app:personalization}.
These spatial and character assets are jointly used later in the first-frame prompting, keyframe review, and video generation.

\textbf{Keyframe-Video Prompt for Clip Generation.}
Given the clip-level script and the scene-level visual assets, we construct a paired keyframe-video prompt for each clip. 
The keyframe prompt specifies the static first frame, including character composition, spatial relations, key prop placement, and camera viewpoint. 
The video prompt describes the temporal development from that starting frame, including character actions, interactions, prop changes, and local narrative progression. 

To improve prompt executability before rendering, we introduce a prompt-level reviewer loop. 
The reviewer checks spatial consistency, physical plausibility, and cross-clip continuity, and further verifies prop continuity across adjacent clips. 
When violations are detected, the system first outputs the issue list, root-cause analysis, and targeted revision suggestions, and then rewrites the corresponding keyframe or video prompt.
In this way, many spatial, physical, and continuity errors can be corrected at the text level before the first frame and video generation.

\subsection{Keyframe-to-Video Generation with 3D Priors}
\label{sec: keyframe to video generation}
Current clip-based video generation pipelines~\cite{movieagent} often synthesize each clip as an independent storyboard shot, or reuse the previous clip's tail frame as the next initial frame. This easily leads to scene drift and struggles to adapt to moving views.
To address these issues, we introduce consistent first-frame synthesis via 3D scene grounding.

\textbf{Scene Anchor Initialization.}
For each scene, we first generate a person-free $360^\circ$ panorama $P$ and reconstruct a scene-level 3D world $\mathcal{W}$ using Marble~\cite{worldlabs2025marble}. 
Since $P$ covers the complete scene, we can sample multiple candidate views from the canonical space. 
Given candidate view parameters $v_1,\ldots,v_K$, we obtain empty background candidates
$
B_k=\Pi(P;v_k), \quad k=1,\ldots,K,
$
where $\Pi$ denotes panorama-to-perspective projection.
For each background $B_k$, an image generation model~\cite{google2026gemini31propreview} synthesizes a character-conditioned first-frame candidate $I_k$ using the background and the scene-level character references. A vision-language model~\cite{qwen3.5} then selects the view that best supports character placement while preserving the scene layout. The selected pair is denoted as $(B^\star,I^\star)$.

\begin{figure}[t]
\centering
\includegraphics[width=0.95\textwidth]{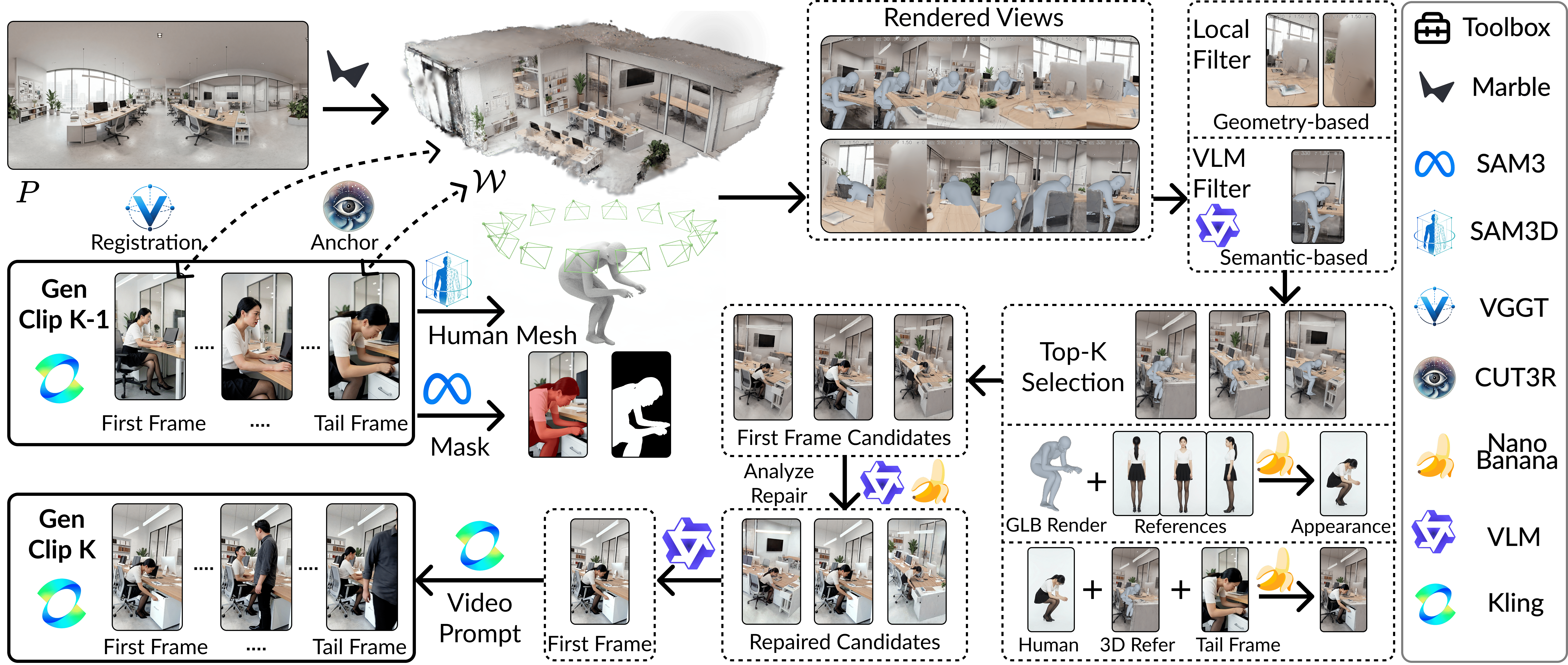}
\vspace{-2mm}
\caption{
Consistent first-frame synthesis via 3D scene grounding. 
We reconstruct a scene-level 3D world from a panorama, register generated clips and the human mesh into the shared coordinate system, and synthesize the next first-frame through geometry-semantic-aware camera selection, character-conditioned generation, and frame quality analysis and repair.
}
\label{3d-consistent-first-frame}
\vspace{-6mm}
\end{figure}
\textbf{First-Frame Registration, Video Trajectory Anchoring, and Human Alignment.}
Although $I^\star$ is generated from the selected background $B^\star$, character insertion may cause small viewpoint or focal-length shifts. We therefore register $I^\star$ back to the 3D world $\mathcal{W}$. Since $B^\star$ is cropped from the panorama, its pose $T_{B^\star}$ and intrinsics are known. After masking the character region, we use VGGT~\cite{wang2025vggt} to estimate the relative transform $\Delta T_{I^\star\rightarrow B^\star}$ and initialize
$
T_{I^\star}=T_{B^\star}\Delta T_{I^\star\rightarrow B^\star}.
$
We resolve the scale ambiguity by aligning the VGGT depth of $B^\star$ with the depth rendered from $\mathcal{W}$, and further refine rotation, translation, and focal length on the background.

After generating the clip from $I^\star$, we anchor its camera trajectory to the same world. We sample video frames and use CUT3R~\cite{wang2025continuous} to recover a local trajectory, depth, and intrinsics. Each frame is expressed relative to the first frame and anchored by
$
T_t=T_{I^\star}\Delta T_t,
$
where $\Delta T_t$ is the scale-calibrated relative pose. We further refine the tail-frame pose $T_{\mathrm{tail}}$ by aligning background-only regions in a local temporal window using color, edge, and depth consistency.
Next, the character is aligned to the shared coordinate system. From the tail frame, SAM 3D Body~\cite{sam3d_body} reconstructs a human mesh with corresponding 2D/3D body keypoints. 
We register the generated mesh to the tail frame based on the keypoints and the person mask from SAM3~\cite{sam3}.

% \yufei{\textbf{Geometry-Adaptive 3D Character Registration.}
% Registering the reconstructed character mesh into the scene-level 3D world $\mathcal{W}$ faces an inherent scale--depth ambiguity: 
% %
% the human 3D mesh is recovered from a single frame, leading to an unknown coordinate transformation.
% %
% The character's mask and 2D keypoints thus fix the character's in-image position and orientation, but not its distance to the camera---a person twice as large and twice as far away produces exactly the same mask. 
% %
% To resolve this, our registration selects a localization strategy according to the geometric cues available in the shot. 
% %
% (i) \textbf{Occlusion matching}: when a scene object occludes the character, the visible silhouette changes as the body moves along the viewing ray, so the body is localized at the depth where the rendered visible silhouette best matches the person mask. 
% %
% (ii) \textbf{Ground-depth reading}: when the character is unoccluded, but its feet are visible and touch the ground, the depth of the ground beneath the feet, read from $\mathcal{W}$, directly determines the character's distance; a standing pose is not required. 
% %
% (iii) \textbf{Floor contact}: for standing characters, the lowest point of the body is placed on the floor of $\mathcal{W}$, which solves its position.
% %
% This places the human, tail-frame camera, and 3D scene in a common coordinate for next-clip planning.}
\textbf{Geometry-Adaptive 3D Character Registration.}
Registering the reconstructed character mesh into the scene-level 3D world $\mathcal{W}$ is challenged by an inherent scale--depth ambiguity: the human mesh is recovered from a single frame and thus has an unknown coordinate transformation.
The character mask and 2D keypoints constrain its image-space position and orientation, but not its distance to the camera---a person twice as large and twice as far away yields the same mask.
We therefore select the registration strategy according to the geometric cues available in the shot.
(i) \textbf{Occlusion matching}: when the character is occluded by a scene object, moving the body along the viewing ray changes the visible silhouette; we localize it at the depth whose rendered silhouette best matches the person mask.
(ii) \textbf{Ground-depth reading}: when the character is unoccluded but its feet are visible and in contact with the ground, the ground depth beneath the feet, read from $\mathcal{W}$, directly determines the character depth, without requiring a standing pose.
(iii) \textbf{Floor contact}: for standing characters, we place the body's lowest point on the floor of $\mathcal{W}$ to determine its position.
This registers the human, tail-frame camera, and 3D scene in a common coordinate system for next-clip planning.

\textbf{Next-Shot Consistent First-Frame Generation.}
The pipeline for next-first-frame and next-clip generation is shown in \cref{3d-consistent-first-frame}. Given the tail-frame pose and aligned human model, we sample geometrically feasible cameras for the next clip on a local spherical shell by varying azimuth, radius multiplier, and elevation. For each camera, we render the background from $\mathcal{W}$, the human mesh, and their rough composite.
The candidates are filtered in two stages. The geometric filter (local filter) removes views that are too close to scene surfaces, heavily occluded in face visibility, or lack sufficient valid background. 
The semantic filter (VLM filter) uses a VLM to verify whether scene anchors from the next-clip prompt are visible in the rendered background. 
% The top-$K$ candidates that satisfy both criteria are kept for the next step.

For each of the top-$K$ selected cameras, we generate a view-conditioned appearance image. The mesh render provides pose, silhouette, and viewpoint constraints, while multi-view character references preserve identity, clothing, and appearance. We then synthesize the next clip's first frame from the rendered reference image with human mesh, generated appearance image, and previous tail frame, so that the layout follows the 3D geometry while identity and cross-shot continuity are preserved.

Finally, a VLM checks the generated first frame for background blur, warped boundaries, missing details, and brightness or color-temperature mismatch. An image generation model repairs the background while preserving the human, camera view, and scene layout, followed by conservative color correction. The resulting first frame is then passed to the frame and video review loop. For clips with multiple characters, we additionally use the nearest previous frame in which other characters are visible to reconstruct their 3D models, and use the center of all involved characters as the camera target; details are provided in \cref{app:multi_character}.

\subsection{Post-Production and Assembly}
\label{sec:post production and assembly}
\textbf{Diverse Scene Transitions.}
Unlike pipelines~\cite{xiaoyunque, movieagent, scriptagent, storymem} that simply concatenate independently generated scenes, ours explicitly designs transitions between adjacent scenes. As shown on the left of \cref{overview}.D, the transition type is selected according to temporal shift, spatial shift, and character movement. If two scenes are continuous in both time and space, we use a direct cut to preserve action immediacy. If the location is unchanged but time advances, we generate a temporal transition with a short text overlay. If the story moves to a substantially different location, we use a location-establishing shot to clarify the upcoming time and place. If the transition involves local spatial movement with narrative meaning, we generate a motion-bridge transition.
%
% , such as a character walking through a corridor.
%
This space-time-aware planning improves scene-to-scene continuity and viewing smoothness without adding unnecessary narrative burden. 

\textbf{BGM Planning and Mixing.}
Since raw audio from video generators may contain artifacts, mismatched music, or inconsistent ambience, we introduce scene-level BGM for emotional continuity. As shown on the right of \cref{overview}.D, we organize a short-drama BGM library of $8{,}122$ tracks into $8$ functional categories, such as dialogue bed, suspense, and conflict escalation, and $40$ fine-grained subcategory buckets using provider metadata including genre, instrument, and speed. For each scene, an LLM selects primary and backup buckets from the scene overview, clip descriptions, clip-level BGM moods, and bucket descriptions. GPT-Audio~\cite{openai2026gptaudio} then scores candidate segments by emotional, narrative, rhythm, and transition fit, and selects the best segment as the scene BGM.
We mix the selected BGM with generated scene audio using adaptive volume control, including dialogue-aware base volume adjustment, LUFS-based loudness calibration, and speech-preserving dynamic compression. This maintains scene-level musical coherence while preserving dialogue clarity. More details are provided in \cref{app:transition_bgm}.

\section{Experiments}
\subsection{Script Corpus and BGM Library Setup}

% \textbf{Short-Drama Script Database}.
To strengthen narrative planning, we build a structured short-drama database from $1{,}102$ original short-drama scripts, which are distilled into $16{,}311$ beat cards and segmented into $4{,}750$ logic chunks. 
We also build a BGM library with $8{,}122$ tracks, covering $8$ high-level categories and $40$ fine-grained subcategories.
More details about the corpus and library can be seen in 
\cref{app:bgm_bank}.
  
\subsection{Experimental Settings} 
\textbf{Short-Drama-Bench.}
Our proposed Short-Drama-Bench consists of $50$ story prompts spanning $7$ popular categories, such as Time Travel \& Rebirth, Suspense \& Thriller, and Sweet Romance, which include $17$ fine-grained subcategories.
Each subcategory contains $2$–$3$ representative samples, covering a broad range of short-drama patterns. More details can be seen in \cref{app:benchmark_samples}.

\textbf{Evaluation Benchmarks and Metrics.} 
To comprehensively evaluate the generated short drama, we utilize three groups of metrics across different benchmarks. First, standard video-generation metrics from VBench~\cite{vbench} are adopted to measure low-level video quality. We also evaluate story visualization ability based on ViStoryBench~\cite{vistorybench}, but adapt its protocol from image-based assessment to multi-frame video evaluation, better fitting our task setting. 

Second, we introduce short-drama-specific metrics for the requirements of our task. For \textit{Narrative Hook}, \textit{Opening Hook} is scored on the first $10$ seconds of the drama and \textit{End Hook} on the ending of each scene. For \textit{Narrative Flow}, \textit{Escalation Effect} is scored on the middle portion of each scene to measure conflict escalation, and \textit{Narrative Coherence} on the complete drama to measure logic clarity.

For \textit{Continuity}, we evaluate adjacent clips by sampling three frames from the previous clip's tail and another three from the next clip's beginning. For \textit{Audio \& Transition}, we evaluate BGM Emotion Alignment at the scene level and transition naturalness. For model-based evaluation, we choose Gemini 3.1 Pro~\cite{google2026gemini31propreview}, Qwen3.5-Omni-Plus~\cite{qwen35omni}, and Seed 2.0 Pro~\cite{seed2026seed20}. 

Third, to complement model-based evaluation, we conduct a human study with $20$ annotators. For each method, we sample the same evaluation units as used in the model-based metrics.
All samples are anonymized and presented in randomized order to reduce method-specific bias.
Annotators rate each sample on a 5-point Likert scale according to the short-drama criterion.
We report the score averaged over all annotators. More details can be seen in \cref{app:Human Rating}.

\textbf{Baselines.} We compare our method with two groups of baselines. The first group includes five story-visualization and long-form video generation pipelines: ScriptAgent~\cite{scriptagent}, MovieAgent~\cite{movieagent}, StoryMem~\cite{storymem}, LongLive~\cite{longlive}, and HoloCine~\cite{holocine}.
Since these methods are not originally designed to take a single-sentence short-drama idea as input, we use Claude Opus 4.6~\cite{anthropic2026claudeopus46} to expand each prompt into the required format of each baseline for a fair comparison.  The second group includes two short-drama products, Toonflow~\cite{toonflow} and Xiao Yun Que~\cite{xiaoyunque}. We use their built-in LLM-based script expansion interface to convert each single-sentence prompt into an executable production script. All baselines are evaluated on the same $50$ prompts in \textit{Short-Drama-Bench}.  Closed-source commercial systems are marked with $^\dagger$ in the tables. More details can be found in \cref{Baseline Setting}.

\subsection{Qualitative Analysis}

\begin{figure}[t]
\centering
\includegraphics[width=1.0\textwidth]{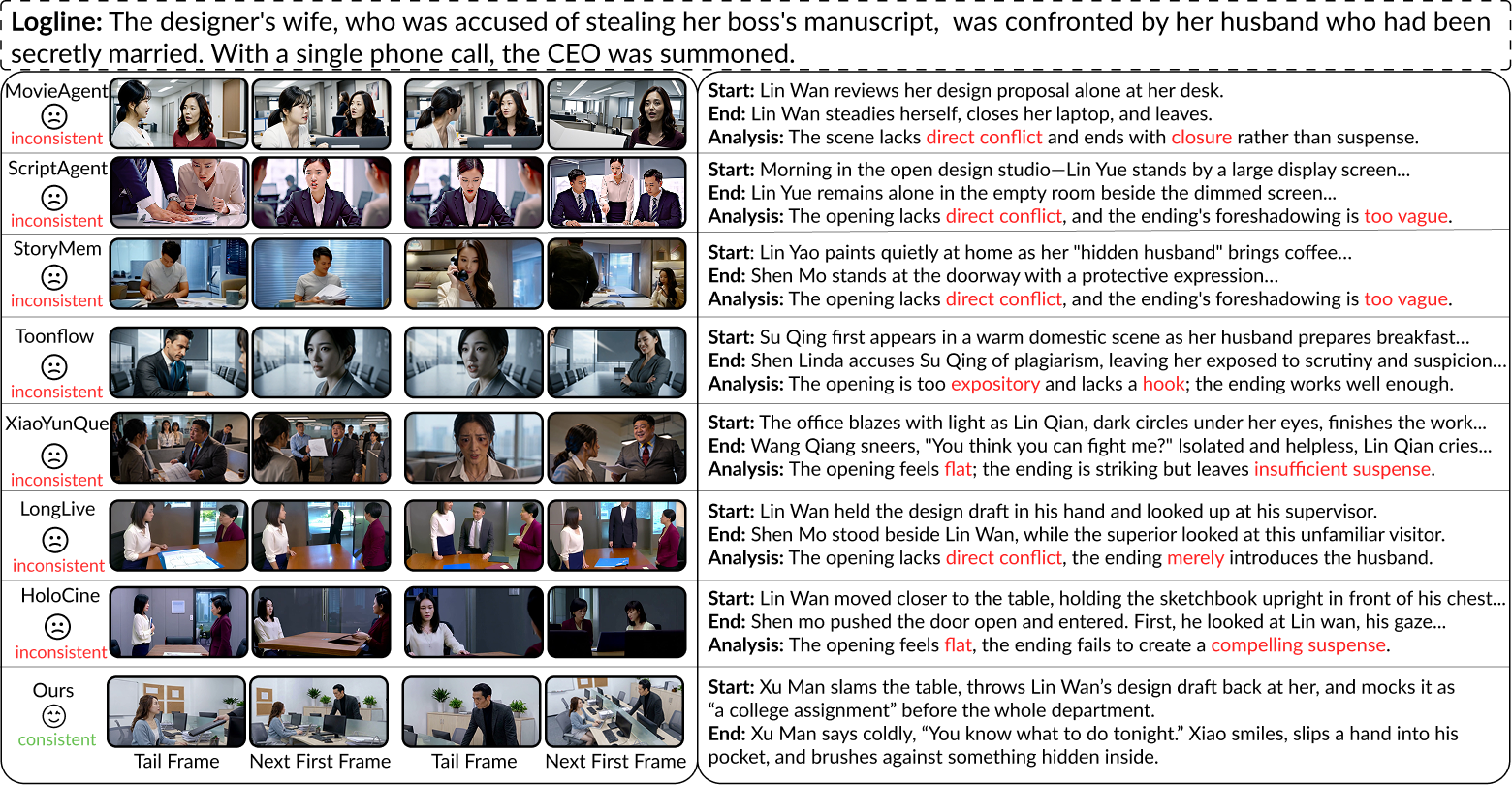}
\vspace{-3mm}

\includegraphics[width=1.0\textwidth]{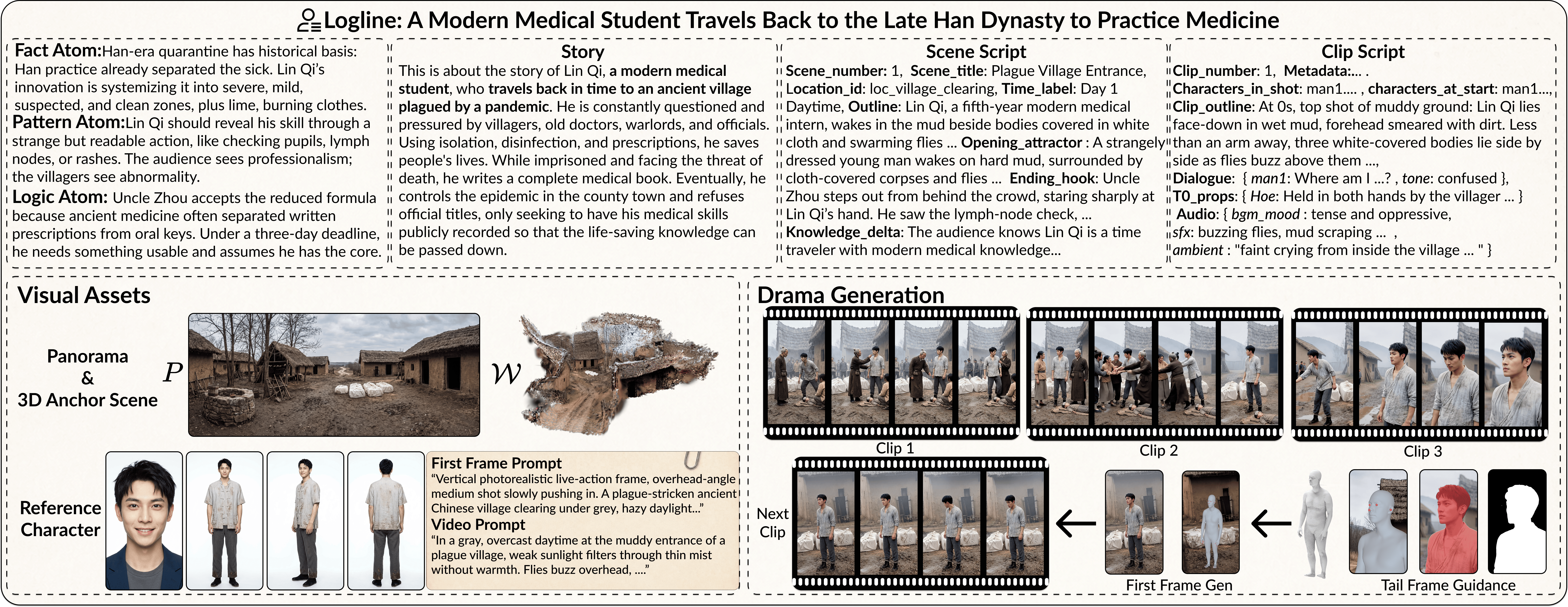}
\vspace{-6mm}

\caption{
Qualitative examples. 
Top: comparison between our generated results and baselines on cross-clip visual continuity and drama pacing. 
Bottom: visualization of our complete drama generation process, including atom retrieval, story structurization, scene/clip script synthesis, visual asset generation, prompt synthesis, and cross-clip 3D-consistent generation.
}
\label{fig:qualitative_combined}
\vspace{-5mm}
\end{figure}

As shown at the top of \cref{fig:qualitative_combined}, we compare representative outputs in terms of spatial continuity and short-drama pace. The left part compares the tail frame of clip $N\!-\!1$ with the first frame of clip $N$. Baselines show noticeable cross-clip drift in both character positions and background layouts. For example, in Xiao Yun Que~\cite{xiaoyunque}, the boss and employee interact across an office partition in the previous tail frame, but move to a corridor-like space in the next first frame. Our method better preserves the scene layout and character blocking by our 3D-grounded first-frame generation mechanism significantly.

The right part of \cref{fig:qualitative_combined} compares generated scripts. Baselines often produce weak openings or scene endings without creating sufficient curiosity. In contrast, our multi-agent debate module strengthens opening conflicts and ending hooks that motivate continued viewing. The examples also show the role of our multi-stage reviewer loops: while commercial platforms such as Toonflow~\cite{toonflow} and Xiao Yun Que~\cite{xiaoyunque} often require manual involvement, our script, image, and video reviewers automatically analyze such errors and trigger targeted revisions. These results illustrate how our framework jointly improves spatial consistency, narrative pacing, and production-level quality control. Moreover, we show a drama's complete script, assets, and clip structure in the bottom part of \cref{fig:qualitative_combined} for better illustration.

\subsection{Quantitative Analysis}

As shown in \cref{tab:quantitative_and_human}, our method achieves strong performance across standard video metrics, story-visualization metrics, and the proposed short-drama-specific metrics. On VBench~\cite{vbench} and ViStoryBench~\cite{vistorybench}, our method improves both general visual quality and story-level consistency, indicating that the proposed framework does not trade off low-level video quality for higher-level narrative control.

On the short-drama-specific metrics, long-form story-visualization baselines such as MovieAgent~\cite{movieagent}, ScriptAgent~\cite{scriptagent}, LongLive~\cite{longlive}, HoloCine~\cite{holocine} and StoryMem~\cite{storymem} generally perform worse on \textit{Opening Hook} and \textit{End Hook}, since they are not explicitly optimized for the compressed pacing and frequent suspense points required by short dramas. Their continuity scores also expose different failure modes. 
MovieAgent and HoloCine generate multiple clips largely from independent textual descriptions, making cross-clip spatial relations difficult to preserve. 
ScriptAgent reuses the previous tail frame as the next clip's first frame, which helps preserve local appearance continuity but limits viewpoint changes. 
StoryMem introduces memory mechanisms to preserve character and scene information across clips, while LongLive uses KV caching and frame-level attention sinks to improve visual consistency in long videos. However, these methods primarily focus on visual consistency and continuous generation, without explicitly addressing short-drama pacing, opening hooks, or end-of-scene suspense.
Moreover, the videos generated by StoryMem, LongLive, and HoloCine are silent, leading to a relatively low \textit{BGM Emotion Alignment} score.
Short-drama production platforms such as Toonflow~\cite{toonflow} and Xiao Yun Que~\cite{xiaoyunque} achieve relatively better narrative and viewing-experience scores, but their visual generation is typically conditioned on a limited set of reference images or visual prompts, leading to missing viewpoints, inconsistent scene layouts, and unstable cross-clip character blocking.
In addition, their BGM mainly relies on the native audio generation capability of the underlying video model, e.g., Seedance~2.0 \cite{seedance2.0}, without explicit scene-level music planning or transition optimization.

\begin{table*}[t]
\centering
\caption{Quantitative evaluations. 
\textbf{Top-left:} comparison on standard video and story-visualization benchmarks.
\textbf{Bottom-left:} comparison on  our proposed Short-Drama-Bench metrics, covering narrative hooks, narrative flow, cross-clip continuity, and audio-transition quality. 
% Scores are averaged over three multimodal judges and 50 benchmark prompts. 
\textbf{Right:} human rating on the same short-drama criteria, averaged over 20 annotators across the benchmark.
The radar plot normalizes each axis by the score of our method to show superior performance over baselines. 
}

\label{tab:quantitative_and_human}

\begin{minipage}[t]{0.64\textwidth}
\centering
\vspace{0pt}

\textbf{General Video and Story Metrics}
\vspace{0.5mm}

\setlength{\tabcolsep}{2.3pt}
\renewcommand{\arraystretch}{0.86}
\resizebox{\linewidth}{!}{%
\begin{tabular}{l ccc cccc}
\toprule
 & \multicolumn{3}{c}{\textbf{VBench}} & \multicolumn{4}{c}{\textbf{ViStoryBench}} \\
\cmidrule(lr){2-4} \cmidrule(lr){5-8}
\textbf{Method} & \textbf{Subject} & \textbf{Background} & \textbf{Motion} & \textbf{Scene} & \textbf{Cross Char.} & \textbf{Global Char.} & \textbf{Single Char.} \\
 & \textbf{Consist.} $\uparrow$ & \textbf{Consist.} $\uparrow$ & \textbf{Smooth.} $\uparrow$ & \textbf{Score} $\uparrow$ & \textbf{Consist.} $\uparrow$ & \textbf{Action} $\uparrow$ & \textbf{Action} $\uparrow$ \\
\midrule
MovieAgent~\cite{movieagent} & 0.8672 & 0.8876 & 0.9084 & 2.5751 & 0.1543 & 2.7883 & 1.6812 \\
ScriptAgent~\cite{scriptagent} & 0.9031 & 0.9092 & 0.9235 & 2.6422 & 0.1926 & 2.3243 & 1.6789 \\
StoryMem~\cite{storymem} & 0.8002 & 0.8719 & 0.9085 & 3.0333 & 0.1753 & 2.4167 & 1.6847 \\
LongLive~\cite{longlive} & 0.9624 & \underline{0.9358} & 0.9711 & 2.9421 & 0.1884 & 2.0922 & 1.9788 \\
HoloCine~\cite{holocine} & 0.9086 & 0.9195 & \underline{0.9741} & \underline{3.4033} & 0.2072 & 2.5367 & 2.6478 \\
Toonflow~\cite{toonflow} & 0.9197 & 0.9303 & 0.9462 & 3.1451 & 0.3405 & 2.2180 & 2.6258 \\
Xiao Yun Que~\cite{xiaoyunque}$^\dagger$ & \underline{0.9216} & 0.8697 & 0.9455 & 3.2527 & \underline{0.4539} & \underline{2.4204} & \underline{2.8928} \\

\midrule
\textbf{Ours} & \textbf{0.9723} & \textbf{0.9465} & \textbf{0.9830} & \textbf{3.5334} & \textbf{0.4606} & \textbf{3.3376} & \textbf{3.0833} \\
\bottomrule
\end{tabular}%
}

\vspace{2mm}

\textbf{Short-Drama-Bench}
\vspace{0.5mm}

\setlength{\tabcolsep}{2.2pt}
\renewcommand{\arraystretch}{0.86}
\resizebox{\linewidth}{!}{%
\begin{tabular}{l cc cc cc cc}
\toprule
 & \multicolumn{2}{c}{\textbf{Narrative Hook}} & \multicolumn{2}{c}{\textbf{Narrative Flow}} & \multicolumn{2}{c}{\textbf{Continuity}} & \multicolumn{2}{c}{\textbf{Audio \& Transition}} \\
\cmidrule(lr){2-3} \cmidrule(lr){4-5} \cmidrule(lr){6-7} \cmidrule(lr){8-9}
\textbf{Method} & \textbf{Opening} & \textbf{End} & \textbf{Escalation} & \textbf{Narrative} & \textbf{Character} & \textbf{Environment} & \textbf{BGM Emotion} & \textbf{Transition} \\
 & \textbf{Hook} $\uparrow$ & \textbf{Hook} $\uparrow$ & \textbf{Effect.} $\uparrow$ & \textbf{Coherence} $\uparrow$ & \textbf{Spatial Cont.} $\uparrow$ & \textbf{Layout Cont.} $\uparrow$ & \textbf{Alignment} $\uparrow$ & \textbf{Naturalness} $\uparrow$ \\
\midrule
MovieAgent~\cite{movieagent} & 2.34 & 1.62 & 1.89 & 1.26 & 2.15 & 1.98 & 2.42 & 1.94 \\
ScriptAgent~\cite{scriptagent} & 1.80 & 2.90 & 2.98 & 2.90 & 2.89 & \underline{3.80} & 2.09 & 2.26 \\
StoryMem~\cite{storymem} & 2.28 & 2.72 & 2.06 & 3.08 & \underline{3.14} & 3.54 & 1.46 & 2.23 \\
LongLive~\cite{longlive} & 1.94 & 2.67 & 2.10 & 2.07 & 2.39 & 2.90 & 1.06 & 1.83 \\
HoloCine~\cite{holocine} & 3.03 & 3.28 & 2.64 & 2.80 & 2.64 & 2.72 & 1.70 & 2.00 \\
Toonflow~\cite{toonflow} & 3.73 & 3.01 & 3.41 & 3.86 & 2.92 & 3.21 & 3.55 & 3.44 \\
Xiao Yun Que~\cite{xiaoyunque} $^\dagger$ & \underline{3.85} & \underline{3.68} & \underline{3.84} & \underline{4.22} & 2.75 & 3.39 & \underline{3.58} & \underline{3.66} \\
\midrule
\textbf{Ours} & \textbf{4.26} & \textbf{3.75} & \textbf{4.06} & \textbf{4.62} & \textbf{3.52} & \textbf{4.05} & \textbf{3.86} & \textbf{3.85} \\
\bottomrule
\end{tabular}%
}
\end{minipage}
\hfill
\begin{minipage}[t]{0.33\textwidth}
\centering
\vspace{0pt}
\textbf{Human Rating}
\vspace{0.5mm}

\includegraphics[width=\linewidth]{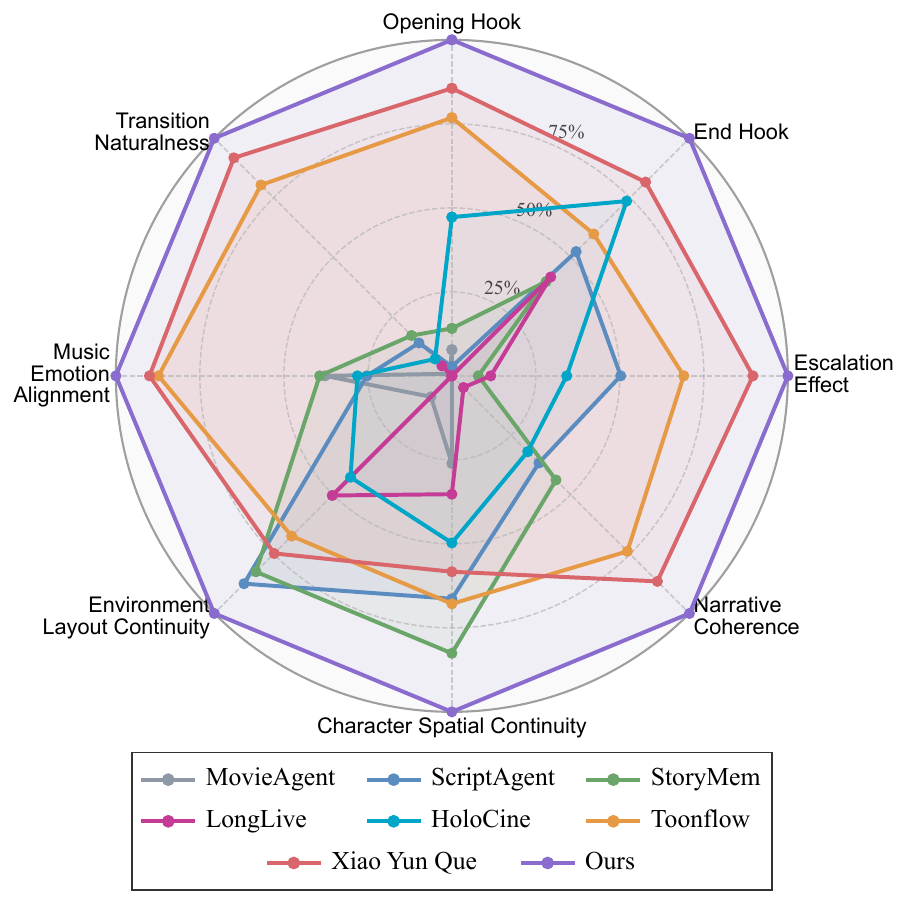}
\end{minipage}
\label{Comprehensive quantitative evaluation}
\vspace{-3mm}
\end{table*}

\vspace{-2mm}
\subsection{Ablation Study}

As shown in \cref{tab:ablation}, we conduct ablations by removing four key components from our full framework. 
Removing Story Gen mainly degrades narrative-related metrics, including opening hooks, ending hooks, and escalation, showing that multi-agent debate is critical for short-drama pacing and suspense construction. 
Removing 3D First-Frame causes the largest drop in continuity metrics, while leaving narrative scores relatively stable, confirming that 3D grounding primarily addresses cross-clip spatial drift. 
Removing Multi-Stage Review leads to consistent degradation across all metrics, indicating that iterative feedback is necessary to correct accumulated errors across script, prompt, keyframe, and video stages.
Removing Transition \& BGM mainly hurts BGM Emotion Alignment and transition naturalness, while clip-level continuity remains nearly unchanged because it is evaluated from adjacent generated clip frames before post-production assembly. 
These results suggest that each component targets a distinct failure mode and jointly contributes to the full system's overall performance.
Isolation ablations of the Idea Bank and the geometry-adaptive registration are reported in \cref{app:isolation_ablation}.

\begin{table*}[h]
\centering
\caption{Ablation study on Short-Drama-Bench. We remove four components from the full system: story generation, 3D first-frame synthesis, multi-stage review, and transition/BGM planning.}
\label{tab:ablation}
\setlength{\tabcolsep}{3pt}
\renewcommand{\arraystretch}{1.0}
\resizebox{\linewidth}{!}{%
\begin{tabular}{l cc cc cc cc}
\toprule
 & \multicolumn{2}{c}{\textbf{Narrative Hook}} & \multicolumn{2}{c}{\textbf{Narrative Flow}} & \multicolumn{2}{c}{\textbf{Continuity}} & \multicolumn{2}{c}{\textbf{Audio \& Transition}} \\
\cmidrule(lr){2-3} \cmidrule(lr){4-5} \cmidrule(lr){6-7} \cmidrule(lr){8-9}
\textbf{Variant} & \textbf{Opening} & \textbf{End} & \textbf{Escalation} & \textbf{Narrative} & \textbf{Character} & \textbf{Environment} & \textbf{BGM Emotion} & \textbf{Transition} \\
 & \textbf{Hook} $\uparrow$ & \textbf{Hook} $\uparrow$ & \textbf{Effect.} $\uparrow$ & \textbf{Coherence} $\uparrow$ & \textbf{Spatial Cont.} $\uparrow$ & \textbf{Layout Cont.} $\uparrow$ & \textbf{Alignment} $\uparrow$ & \textbf{Naturalness} $\uparrow$ \\
\midrule
w/o Story Gen           & 3.48 & 3.02 & 3.31 & 3.82 & 3.34 & 3.84 & 3.70& 3.67 \\
w/o 3D First-Frame      & 4.26 & 3.60 & 3.92 & 4.51 & 2.81 & 3.02 & 3.85 & 3.77 \\
w/o Multi-Stage Review  & 3.86 & 3.32 & 3.66 & 4.28 & 3.04 & 3.48 & 3.47 & 3.42 \\
w/o Transition \& BGM   & 4.22 & 3.72 & 4.02 & 4.49 & 3.52 & 4.05 & 3.15 & 3.03 \\
\midrule
\textbf{Full (Ours)}    & \textbf{4.26} & \textbf{3.75} & \textbf{4.06} & \textbf{4.62} & \textbf{3.52} & \textbf{4.05} & \textbf{3.86} & \textbf{3.85} \\
\bottomrule
\end{tabular}%
}
\vspace{-3mm}

\end{table*}

\FloatBarrier
\vspace{-2mm}
\section{Conclusion}
We presented \textit{One Sentence, One Drama}, a hierarchical multi-agent framework for generating complete personalized short dramas from a single-sentence idea. Our framework addresses three central challenges in this setting: short-drama pacing, cross-clip spatial consistency, and production-level quality control. It combines retrieval-augmented multi-agent story generation, 3D-grounded first-frame synthesis, multi-stage reviewer loops, and scene-level transition and BGM planning. We further introduced \textit{Short-Drama-Bench}, a benchmark with short-drama-specific evaluation criteria. Experiments across automatic metrics, adapted story-visualization evaluation, and human ratings show that our method improves narrative engagement, visual continuity, and overall viewing experience over existing pipelines. These results suggest that structured agentic generation is a promising direction for controllable long-horizon video creation.

\subsection*{AI use statement}
We used generative AI tools to assist in implementing parts of the pipeline code and to polish the English writing (grammar, spelling, and word choice). We did not use generative AI tools to develop the method, design experiments, or create figures. All AI-assisted code was verified and tested by the authors, all edited text was reviewed by the authors, and we take full responsibility for the content of this work.

\subsection*{Ethics statement}
The script corpus used for retrieval is released under the MIT License.
The BGM library contains only royalty-free or commercially usable tracks (\cref{Limitation}).
As with any video generation system, our framework could be misused to produce harmful or misleading content; the safeguards implemented in the released system are described in \cref{app:broader_impacts}.

\subsection*{Reproducibility statement}
The full implementation, prompts, reviewer schemas, and evaluation scripts are provided in the supplementary material and at \url{https://anonymous.4open.science/r/one_sentence_one_drama_ICLR/}. Benchmark prompts, model versions, baseline settings, and hardware are documented in \cref{app:benchmark_samples}, \cref{Baseline Setting}, and \cref{app:review_metrics}.

% (This section is \textbf{recommended} and does not count toward the page limit.)

% It is important that the work published in ICLR is reproducible. Authors are
% strongly encouraged to include a paragraph-long Reproducibility Statement at the
% end of the main text (before references) to discuss the efforts that have been
% made to ensure reproducibility. This paragraph should not itself describe
% details needed for reproducing the results, but rather reference the parts of
% the main paper, appendix, and supplemental materials that will help with
% reproducibility. For example, for novel models or algorithms, a link to an
% anonymous downloadable source code can be submitted as supplementary materials;
% for theoretical results, clear explanations of any assumptions and a complete
% proof of the claims can be included in the appendix; for any datasets used in
% the experiments, a complete description of the data processing steps can be
% provided in the supplementary materials. Each of the above are examples of
% things that can be referenced in the reproducibility statement.

% \clearpage
% \newpage

\bibliographystyle{plainnat}

\bibliography{ref}
% \section{Submission of papers to NeurIPS 2026}
\newpage
\appendix

\section*{Appendix Overview}
\addcontentsline{toc}{section}{Appendix Overview}

This appendix provides additional details for the related work, generation pipeline, benchmark construction, evaluation protocol, implementation settings, prompts, and responsible-use discussion.

\begin{itemize}
    \item \textbf{Appendix A: Broader Impacts.} Potential positive impacts on creative access and production cost, as well as copyright and licensing concerns.
    \item \textbf{Appendix B: Related Work.} Extended discussion of video generation, story visualization, and short-drama generation.
    \item \textbf{Appendix C: Multi-Agent Debating-Based Story Generation.} Details of atom script corpus construction, problem-driven retrieval, story drafting, and debate-based polishing.
    \item \textbf{Appendix D: Diverse Transition Clips and BGM Planning.} Details of scene transition design, BGM subcategory buckets selection, audio scoring, and adaptive mixing.
    \item \textbf{Appendix E: Multi-Character 3D-Consistent First-Frame Generation.} Extension of the 3D-grounded first-frame pipeline to multi-character clips.
    \item \textbf{Appendix F: Human Rating Detail.} Human rating protocol, anonymization, randomization, and score aggregation.
    \item \textbf{Appendix G: Detailed Experimental Settings.} Hardware settings, baseline execution environment, retry policy, and 3D candidate selection.
    \item \textbf{Appendix H: Time and Cost Analysis.} Wall-clock runtime estimates and API cost analysis.
      \item \textbf{Appendix I: Limitations.} Practical limitations including cost, human-in-the-loop interaction, and audio licensing.
    \item \textbf{Appendix J: Multi-Stage Review Metrics and Judge Models.} Internal reviewer metrics, judge models, decision rules, and external benchmark evaluation settings.
     \item \textbf{Appendix K: Reliability and Sensitivity Analysis.} Human--LLM evaluation agreement, robustness to LLM judge choice, statistical confidence of performance improvements, and human-evaluation reliability.
     \item \textbf{Appendix L: Isolation Ablations of the Idea Bank and Geometry-Adaptive Registration.} 
    \item \textbf{Appendix M: Short-Drama-Bench Prompts and Generated Videos.} Benchmark categories, prompt topics, and representative generated video examples.
    \item \textbf{Appendix N: Script Library and BGM Library Details.} Statistics and construction details of the script retrieval corpus and BGM library.
    \item \textbf{Appendix O: Prompt Templates for Each Stage.} Prompt templates for evaluation, text review, image review, and video review.
    \item \textbf{Appendix P: Structures and Examples of Narrative Memory Units.} Exact structures and representative examples of BeatCards, Pattern Atoms, Logic Chunks, and Logic Atoms.
    \item \textbf{Appendix Q: Personalization from user-provided portraits.} 
    
    \item \textbf{Supplementary Material Files.}
    The submitted supplementary package contains the code and a demo short-drama video showcase, including one Chinese-dubbed and one English-dubbed generated video.
\end{itemize}

\newpage

\section{Broader Impacts}
\label[appendix]{app:broader_impacts}

Our framework may broaden access to short-drama creation by reducing the gap between a human creative idea and a complete audio-visual production. 
By turning a single-sentence concept into scripts, visual assets, coherent video clips, transitions, and BGM, the system can lower production cost and technical barriers for independent creators, educators, small studios, and users without professional filmmaking resources. 
It may also support faster prototyping of narrative ideas, multilingual short-drama production, and more diverse forms of personalized storytelling.

% At the same time, automated short-drama generation may raise copyright and licensing concerns, especially when generated stories, visual styles, voices, or music resemble protected works or commercial assets. Practical deployment should therefore use copyright-aware training and retrieval sources, licensed audio-visual assets, and clear policies for generated-content ownership and attribution.

\textbf{Safety measures in the released system.}
The released implementation already contains three safeguards, all of which are verifiable in the code.

(i) \emph{Script-level content constraints.} The script generation prompt (\texttt{step1\_story\_init\_full\_script.py}) explicitly restricts sensitive visual content: when death, injury, violence, blood, remains, or burning are involved, only their narrative function is kept and expressed in a restrained, non-gory manner suitable for image and video generation; elaboration into detailed horror imagery is prohibited.

(ii) \emph{Dialogue-level term replacement.} During dialogue polishing (\texttt{step1\_1\_clip\_dialogue\_role\_polish.py}), a built-in list of about $30$ high-risk terms (e.g., corpse, gore, bloody, rotting, and their Chinese counterparts) triggers a local replacement patch by the polishing LLM. Only the flagged words are rewritten while the plot is locked---dialogue structure, events, conflict intensity, and prop positions remain unchanged---and the patched text is required to contain no remaining risk terms; for example, ``a rotting corpse'' becomes ``the outline of the deceased under a white sheet''.

(iii) \emph{Generator-side filtering.} All images and videos are produced by commercial generators with built-in content filtering (Gemini 3.1 Pro Image Preview~\cite{google2026gemini3proimagepreview} and Kling~\cite{kling}).
These measures complement the multi-stage reviewer loops, which can be extended with explicit safety criteria without changing the architecture.

\section{Related Work}
\label[appendix]{app:related work}
\subsection{Video Generation}
\vspace{-1mm}
% Recent video foundation models have rapidly advanced text- and image-to-video generation in visual fidelity, motion realism, and prompt adherence. Representative works include closed-source systems such as Sora~\cite{sora}, Veo~\cite{veo3}, Kling~\cite{kling}, and Seedance~\cite{seedance2.0}, as well as open-source counterparts such as CogVideoX~\cite{yang2024cogvideox}, HunyuanVideo~\cite{kong2024hunyuanvideo}, and Wan~\cite{wan2025wan}. However, these models are typically limited to $5$–$15$ seconds per clip, far short of the multi-minute, multi-shot requirement of short dramas. They thus serve as per-shot rendering backends in our framework, but cannot by themselves ensure long-horizon planning or cross-clip consistency.
Recent video foundation models have rapidly advanced text- and image-to-video generation in visual fidelity, motion realism, and prompt adherence. Representative works include closed-source systems such as Sora~\cite{sora}, Veo~\cite{veo3}, Kling~\cite{kling}, and Seedance~\cite{seedance2.0}, as well as open-source counterparts such as CogVideoX~\cite{yang2024cogvideox}, HunyuanVideo~\cite{kong2024hunyuanvideo}, and Wan~\cite{wan2025wan}. However, these models are typically limited to $5$–$15$ seconds per clip. To go beyond this limit, autoregressive long-video methods~\cite{causvid,selfforcing,selfforcing++,causalforcing,resampling-forcing,relaxforcing,rollingsink,longlive,skyreels2} roll out clips sequentially with teacher- or self-forcing, scaling generation to the 1-minute level. Nevertheless, both families do not model multi-shot pacing, cross-scene spatial layout, or production-level coherence, and even 1-minute level outputs fall short of the multi-minute, multi-scene requirement of short dramas. They thus serve as per-shot rendering backends in our framework, but cannot by themselves ensure long-horizon planning or cross-clip consistency.

Joint audio-video generation further introduces modality-specific fidelity and synchronization requirements. AV-GRPO~\cite{xu2026av} uses modality-anchored reinforcement learning to improve generation quality, semantic alignment, and cross-modal synchronization. These clip-level objectives complement scene-level music planning and cross-scene audio assembly.
\vspace{-2mm}

\subsection{Story Visualization}
\vspace{-1mm}
To extend single-clip models towards narrative video, prior work explores LLM-guided planning, memory-conditioned generation, and multi-agent collaboration. Early efforts such as VideoDirectorGPT~\cite{lin2023videodirectorgpt}, VideoStudio~\cite{long2024videostudio}, and StoryDiffusion~\cite{zhou2024storydiffusion} use layout planning or shared self-attention to improve cross-scene consistency, while VideoGen-of-Thought~\cite{zheng2024videogen}, StoryAgent~\cite{hu2024storyagent}, MovieAgent~\cite{movieagent}, and ScriptAgent~\cite{scriptagent} adopt multi-agent or chain-of-thought decomposition to organize scripts, storyboards, and shots. StoryMem~\cite{storymem} further reformulates multi-shot generation as iterative synthesis conditioned on a visual memory bank. Despite these advances, these systems are primarily designed for general storytelling or movie-style narratives rather than short dramas. Most of them require carefully curated inputs. For example, MovieAgent~\cite{movieagent} needs a full script and a character bank with reference portraits~\cite{movieagent}, and StoryMem~\cite{storymem} expects detailed per-shot prompts~\cite{storymem}. Many also rely on local fine-tuning of the image or video modules~\cite{movieagent,storymem,hu2024storyagent}, which shifts much of the creative burden to the user. Their shots are typically produced as loosely coupled five-second clips~\cite{movieagent,scriptagent,hu2024storyagent}, with audio either absent~\cite{storymem} or added post-hoc, leading to visible disconnection between adjacent shots. 
Moreover, these systems do not explicitly model dense hooks, frequent reversals, and compressed payoff structures, and thus tend to produce unsatisfactory pacing.

Iterative feedback also extends beyond story synthesis. SciEducator~\cite{xu2026scieducator} uses a Deming-cycle multi-agent loop for scientific video understanding and education. Our framework applies structured review to the narrative and production constraints of short dramas.
\vspace{-2mm}

\subsection{Short-Drama Generation}
\vspace{-1mm}
A few recent systems specifically target short-drama production. Toonflow~\cite{toonflow} is an open-source workflow that converts a full novel into a short drama through sequential character extraction, script generation, storyboard drawing, and video synthesis, while Xiaoyunque~\cite{xiaoyunque} is a closed-source commercial product built on Seedance $2.0$ \cite{seedance2.0}. Despite their popularity, both systems share several limitations. On the script side, Toonflow \cite{toonflow} requires a complete novel as input, while Xiaoyunque \cite{xiaoyunque} appears to rely on one-shot LLM expansion, leading to weak hooks and brittle narrative logic. On the visual side, keyframes are generated independently from a few reference images, causing spatial drift and inconsistent character placement across clips. They also depend on manual inspection for quality control, and neither models scene-level audio or transitions, typically reusing the video model's built-in audio and concatenating scenes via hard cuts. In contrast, our framework addresses these issues through retrieval-augmented multi-agent story generation, 3D-grounded synthesis, multi-stage reviewer loops, and scene-level BGM matching with space-time-aware transition planning.

\subsection{Geometry-Grounded Visual Control}
\vspace{-1mm}
Explicit geometry provides spatial control beyond textual prompts and visual memories. Proxy2World~\cite{xu2026proxy2world} controls scene layout and motion through lightweight proxies, while LaS-Comp~\cite{yan2026comp} completes partial 3D observations using generative shape priors, and CNS-Edit++~\cite{hu2026cns} supports local 3D shape manipulation with a coupled neural representation. These works address scene generation, object completion, and shape editing, respectively. At the image level, Deep Homography Mixture~\cite{yan2023deep} models camera-induced rolling-shutter deformation through learned motion bases for single-image correction. Our framework uses a shared scene coordinate system to preserve character placement and camera relations across shots, with geometry recovered using VGGT~\cite{wang2025vggt} and CUT3R~\cite{wang2025continuous}.
\vspace{-2mm}

\subsection{Multimodal Understanding and Feedback}
\vspace{-1mm}
Reviewing a personalized drama requires tracking who acts, what changes, and whether textual descriptions agree with visible evidence. PVChat~\cite{shi2025pvchat} studies subject-aware video understanding from a single reference video. For dynamic point clouds, 4DPC$^2$hat~\cite{zhang20264dpc} combines temporal reasoning with failure-aware bootstrapping that generates targeted supervision. Actor as Its Own Critic~\cite{zhang2026actor} instead links region understanding and localization through a region--text--region cycle-consistency reward. These methods address complementary understanding and training tasks; our reviewer loops use pretrained models at inference to diagnose production errors and trigger targeted revisions.
\vspace{-2mm}

\section{Details of the Multi-Agent Debating-based Story Generation Framework}
\subsection{Atom Script Corpus Construction} 
% 直接通过一句话来扩写整个短剧剧本，经常会遇到两个大问题，一个是缺乏短剧的节奏，例如开场不够吸引人，结尾悬念不足等问题，一个是叙事因果链不稳定人物为什么这样做说不清，证据为什么此时生效说不清，上一场结尾和下一场开头接不住，剧情容易显得硬拐。，我们这里收集了200个爆火的实际的短剧剧本，首先提取为200个structured script cards，包含每一个剧本的title，剧情，hook，然后从中提炼出来了可以复用的3000个可以重复使用的beat card，每一个都包含 一些基本信息，opening_action，closing_hook_visual，并且通过text encoder变成了embedding，便于后续的检索，例如一个“公开羞辱”型开场，例如一个“身份揭示”型结尾钩子，便于新的剧本生成借鉴pattern，快速掌握短剧节奏，另一方面，我们将原文切分成了多个重叠的文本块，

 As shown in \cref{Appendix multi-agent detail} (Retrieval Bank Construction), directly expanding a full short-drama script from a single logline often suffers from two complementary failure modes: weak short-drama pacing, such as unconvincing openings and underpowered ending hooks, and unstable local causal coherence, where character actions are under-motivated, evidence becomes effective at unclear moments, and scene-to-scene consequences fail to connect smoothly. To address these issues, we construct two complementary retrieval banks from a corpus of $1{,}102$ high-performing short-drama scripts released under the MIT License. First, we distill each source script into a structured script card containing script-level metadata, plot summaries, and further decompose these cards into  $16{,}311$ reusable beat-level units. Each beat unit encodes key structural cues such as the opening action, beat summary, and closing hook visual, and is mapped into an embedding space to support retrieval of transferable short-drama patterns. This forms our Pattern Bank, which provides reusable pacing and packaging priors for new stories. Second, we segment the original scripts into overlapping local text chunks to preserve short-range causal context across boundaries. These chunks constitute a Logic Bank that supports retrieval of local narrative evidence, including motivation chains, evidence activation conditions, and consequence transitions. 

\subsection{Problem-Driven Retrieval}
% 如图1的 Problem-Driven Retrieval 部分所示，在仅给定一条 logline 的条件下，直接扩展出完整短剧剧本往往缺乏足够的外部支撑。为此，我们首先利用大语言模型将 logline 扩展为包含初步剧情骨架与关键冲突线索的 seed text，并进一步分析当前故事生成所缺失的支撑信息，从而生成一个面向问题的检索计划。该计划包含三类互补的检索需求。第一，对于医学、历史、法律、制度流程等具有外部事实约束的内容，我们触发基于 Web search 的事实检索，以补充专业知识和现实规则。第二，对于人物动机、证据生效条件、场间后果承接以及信息揭示顺序等局部叙事成立性问题，我们到 Logic Bank 中检索相关的局部文本块，并选取最相关的 top-k 结果作为局部因果支撑。第三，对于开场设计、冲突包装、反转节奏和结尾钩子等短剧化表达需求，我们在 Pattern Bank 中对 beat-level card 的多个视角表示分别计算相似度，并进行加权融合排序，从而检索出最相关的 top-k 短剧模式参考。最后，我们将三路检索结果统一送入一个摘要模块，将原始检索内容压缩为可复用的结构化单元，即 Fact Atoms、Logic Atoms 和 Pattern Atoms。这一过程一方面为后续剧本规划提供事实、因果与节奏层面的互补先验，另一方面也避免了对原始剧本文本的直接复制，使检索结果以抽象、可迁移的形式服务于新剧本生成。

As shown in \cref{Appendix multi-agent detail}, directly expanding a complete short-drama script from a single logline often lacks the external support needed for coherent and compelling story planning. To address this, we first use an LLM to expand the input logline into a seed text containing a preliminary narrative skeleton and key conflict cues. Based on this seed text, the LLM further analyzes what kinds of support are missing and generates a structured retrieval plan with three complementary retrieval routes. First, for externally grounded content such as professional knowledge, historical details, legal constraints, and institutional procedures, we invoke web search to retrieve factual evidence. Second, for local narrative validity, including character motivation, evidence activation conditions, scene-to-scene consequence chaining, and knowledge-state transitions, we retrieve the top-k relevant local chunks from the Logic Bank as causal support. Third, for short-drama-specific dramatic packaging, including opening design, conflict presentation, reversal pacing, and ending hooks, we retrieve the top-k most relevant beat-level cards from the Pattern Bank by computing similarity over multiple beat views and aggregating them with weighted ranking. Finally, we feed all retrieved evidence into a summarization module, which compresses the raw retrieval outputs into reusable structured units, namely Fact Atoms, Logic Atoms, and Pattern Atoms. This process provides complementary factual, causal, and pacing priors for downstream story planning, while also avoiding direct copying of source scripts by transforming retrieved content into abstract, transferable units.

\subsection{Story Drafting}
 % 在获得 logline、seed text 以及检索得到的事实、逻辑和短剧模式支持后，我们进入正式的故事扩写阶段。该阶段首先生成一个 story core，其中包含故事级别的元信息以及逐场的 scene plan。具体而言，story core 不仅定义标题、主题、类型和整体叙事框架，还为每一场戏生成结构化规划，包括场景标题、时空边界、剧情概要、开场吸引
 %  点、关键推进步骤、场次目标、中段升级点以及结尾钩子。为了同时强化全局因果约束、场间承接和短剧化节奏控制，我们进一步引入五条跨场景的全局推进线来统领整部故事的发展，即外部压力线、主角应对线、定局机制线、情绪变化线和知识状态线。前四条线分别约束故事中的压迫升级、主角策略与资源变化、最终解决机制的铺垫，以及情
 %  绪张力的动态起伏；知识状态线则显式记录每一场戏之后观众和剧中角色分别知道什么、尚不知道什么，以及新增了哪些关键信息或证据，从而减少信息混乱和叙事断裂。在 story core 生成之后，我们进一步扩展得到 story assets，用于为后续场景扩写和视频生成提供一致的结构化资产支持。story assets 主要包含三类内容：角色资产、空间资产和道具资产。具体而言，我们为主要人物生成稳定的身份与外观描述，为核心场景生成可复用的空间描述与视觉属性，并为关键道具生成功能、象征意义和出现约束等信息。
 
   After obtaining the logline, the seed text, and the retrieved factual, logical, and pattern-level support, we perform story drafting in two stages. We first construct a story core, which specifies both story-level metadata and a structured scene plan for the entire drama. Concretely, the story core defines the title, theme, genre, and overall narrative framing, and, for each scene, predicts the scene title, spatiotemporal boundary, outline, opening attractor, key progression steps, scene goal, escalation beats, and ending hook. To maintain global consistency beyond isolated scene planning, we introduce five cross-scene progression lines: the external pressure line, the protagonist response line, the resolution mechanism line, the emotional progression line, and the knowledge-state line. The first four lines organize the escalation of external constraints, the evolution of the protagonist’s strategy and resources, the gradual setup of the eventual resolution, and the trajectory of emotional tension, respectively. The knowledge-state line records, after each scene, what the audience and the in-story characters know, what remains hidden, and which new evidence or state changes have been introduced, thereby improving information control and scene-to-scene coherence. We then derive story assets from the resulting story core, including character assets, location assets, and prop assets. These assets provide stable identity and appearance descriptions for major characters, reusable spatial descriptions and visual attributes for core locations, and functional as well as symbolic descriptions for key props.

\subsection{Multi-Agent Debate Polishing}
\label[appendix]{app:multi_agent_debate}
\begin{figure}[t]
\centering
\includegraphics[width=1.0\textwidth]{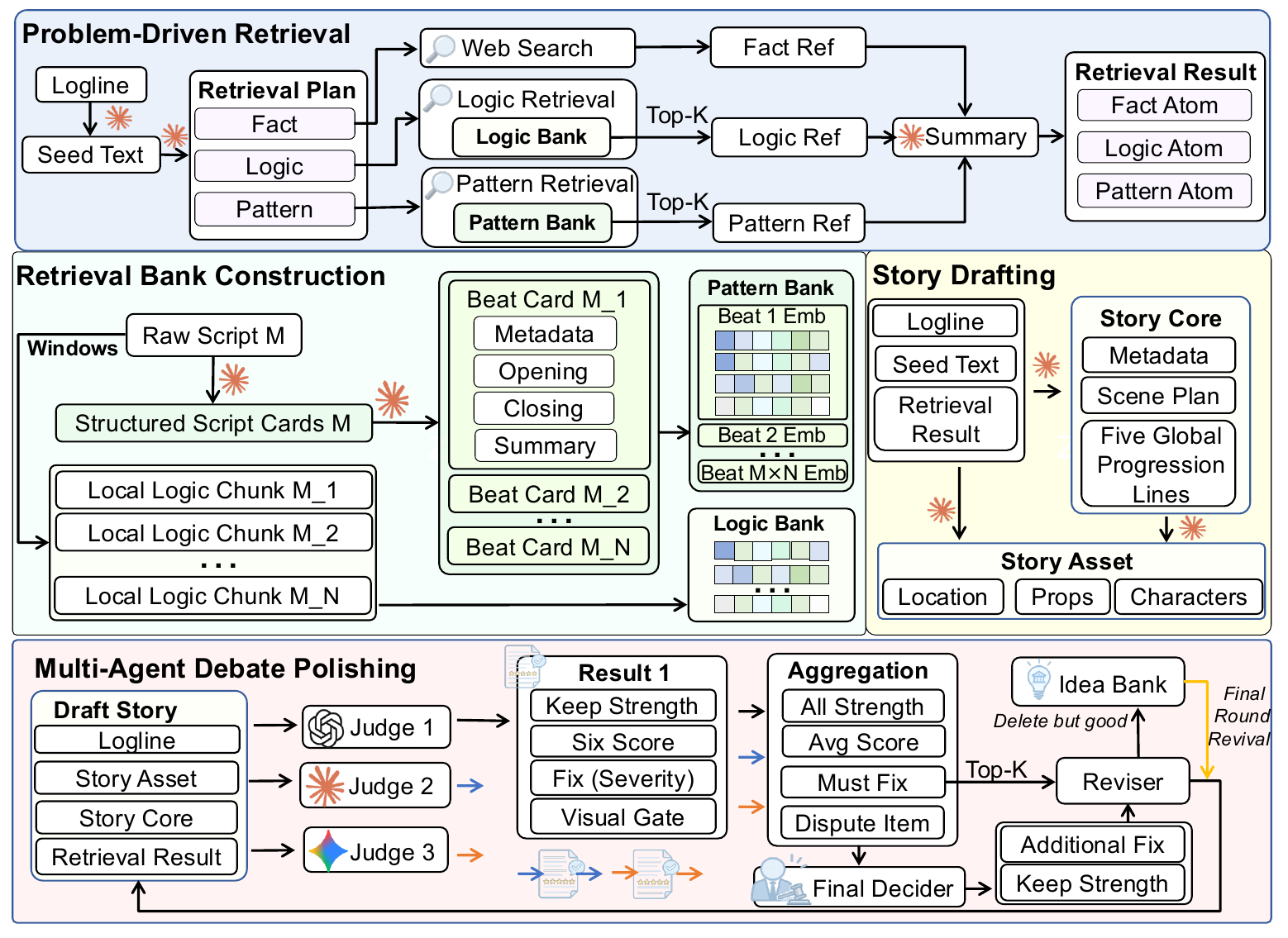}
\vspace{-4mm}
\caption{The Multi-Agent Debating-based Story Generation Framework.}
\label{Appendix multi-agent detail}
\vspace{-4mm}
\end{figure}

% 在 Multi-Agent Debate Polishing 阶段，我们将初始生成的 Draft Story 同时送入三个彼此独立的前沿大语言模型 Judge 中进行并行评审。每个 Judge 都基于同一份输入，返回一组结构化评审结果，包括需要保留的亮点 (keep strengths)、六个维度的评分、必须修复的问题以及视觉可执行性判断。其中，六个评分维度分别为：逻辑完整性、开场吸引力、钩子承接性、叙事清晰度、反转节奏以及兑现与收束效果。对于每个必须修复的问题，Judge 还会给出严重程度、问题证据、修复方向以及对应的修改目标。除分数外，每个 Judge 还会给出一个 visual executability gate，用于判断当前草稿中的关键转折是否能够被后续场景扩写与视频生成有效落地。在获得三个独立评审结果后，我们采用一个确定性的 aggregation 过程对其进行统计、合并与去重，得到全局保留亮点、各维度平均分、候选 must-fix 列表以及争议项集合。争议项主要包括以下几类情况：不同 Judge 在同一维度上的评分差异过大、对同一问题给出的严重程度差异显著、不同 Judge 提出的修复方向明显冲突，或存在逻辑完整性过低、视觉可执行性失败等高风险信号。对于这些争议项，我们进一步引入一个 Final Decider 进行选择性裁决。在实现上，Final Decider 由能力最强的模型承担，用于判断争议问题是否需要修复、应遵循何种最小修改原则，以及哪些高价值亮点必须受到保护。随后，我们将 aggregation 选出的 top-k 个 must-fix 问题与 Final Decider给出的裁决结果共同送入 Reviser，由其执行基于 patch 的局部修订，而不是整稿重写。在修订过程中，凡是为了增强逻辑稳定性、清晰度或场间承接而被删除或削弱的优秀创意、钩子设计或记忆点，都会被显式记录到 Idea Bank 中。每轮修订完成后，新的草稿会再次进入相同的多模型评审与聚合流程；若评分达到预设阈值，则直接停止，否则继续进入下一轮修订，最多迭代 (N) 轮。在最后一轮完成后，我们还会执行一次 final-round revival：在不破坏当前逻辑完整性与可执行性的前提下，从 Idea Bank 中选择少量可安全恢复的高价值创意重新注入故事，从而尽量减少修订过程中对开场吸引力、悬念强度和记忆点的过度损失。
In the Multi-Agent Debate Polishing stage, we submit the drafted story to three independent frontier LLM judges for parallel review. Given the same input, each judge returns a structured evaluation that includes keep strengths, six rubric scores, must-fix issues with severity levels, and a visual executability gate. The six scoring dimensions are logical integrity, opening strength, hook continuity, narrative clarity, reversal pacing, and payoff resolution. For each must-fix issue, the judge further specifies the supporting evidence, the recommended fix direction, and the target object that should be revised. The visual executability gate provides an additional non-score signal indicating whether key turning points can be reliably grounded in downstream scene expansion and video generation.
 
We then perform deterministic aggregation over the three reviews to merge, deduplicate, and summarize the outputs into a unified set of retained strengths, average rubric scores, candidate must-fix issues, and disputed items. Disputed items arise when judges exhibit large score discrepancies on the same dimension, assign substantially different severities to the same issue, propose conflicting fix directions, or expose high-risk signals such as failed visual-executability checks or critically low logical-integrity scores. These disputed items are further routed to a Final Decider, implemented with GPT-5.4 Pro, which selectively determines whether a disputed issue should be fixed, what minimal-change principle should be followed, and which strengths must be explicitly protected. We then pass the top-k aggregated must-fix issues together with the decider’s rulings to a Reviser. Rather than regenerating the entire draft, the Reviser performs patch-based local revision: it outputs structured patches that replace only the targeted scene plans, or, when necessary, the global cross-scene progression lines. For each revised scene, the patch rewrites the scene outline together with its dependent fields, including the opening attractor, key progression steps, scene goal, escalation beats, ending hook, and knowledge-state update, thereby enforcing minimal but coherent modifications while preserving unaffected parts of the story.
  
During revision, any strong idea, hook, or memorable dramatic design that is removed or softened in order to improve logic, clarity, or continuity is explicitly recorded in an Idea Bank. After each revision round, the updated draft re-enters the same multi-judge review and aggregation loop. The process terminates once the draft satisfies predefined quality thresholds, or after at most $3$ rounds. Finally, we perform a final-round revival step, which revisits the Idea Bank and selectively restores a small number of previously removed ideas when they can be reintroduced without harming the current logical integrity or executability. This final step mitigates over-correction during iterative polishing and helps recover strong hooks, payoffs, and memorable dramatic moments.

\section{Details of Diverse Transition Clips and BGM Planning \& Mixing}

\subsection{Diverse Transition through Scenes}
\label{sec:clip transition}

Unlike conventional long-video generation pipelines that simply concatenate independently generated scenes, we explicitly model the transitional relationship between adjacent scenes. Hard cuts are easy to implement, but they often introduce two problems in long-form narratives: visually abrupt pacing, which weakens the viewing experience, and ambiguous temporal or spatial context, which makes it difficult for viewers to infer when and where the next scene takes place. As shown in \cref{bgm_transition}, we introduce diverse transition clips that are selected according to the temporal shift, spatial shift, and character movement between consecutive scenes. When two scenes are continuous in both time and space, we use a direct cut to preserve the immediacy of the action. When the location remains largely unchanged but time advances substantially, we generate a temporal transition, such as a time-lapse exterior
shot of the office building, with a short text overlay indicating the elapsed time. When the story moves to a substantially different location, we generate a location-establishing transition, using an exterior or entrance shot of the next location together with a text overlay that clarifies the upcoming time and place. When the transition involves only a local spatial change and the character movement itself carries narrative information, we generate a motion-bridge transition, such as a character walking through a corridor or moving toward an elevator, to visually connect the two scenes. This space-time-aware transition planning improves scene-to-scene continuity, interpretability, and viewing smoothness without adding unnecessary narrative burden.

\subsection{BGM Planning \& Mixing}
\begin{figure}[t]
\centering
\includegraphics[width=1.0\textwidth]{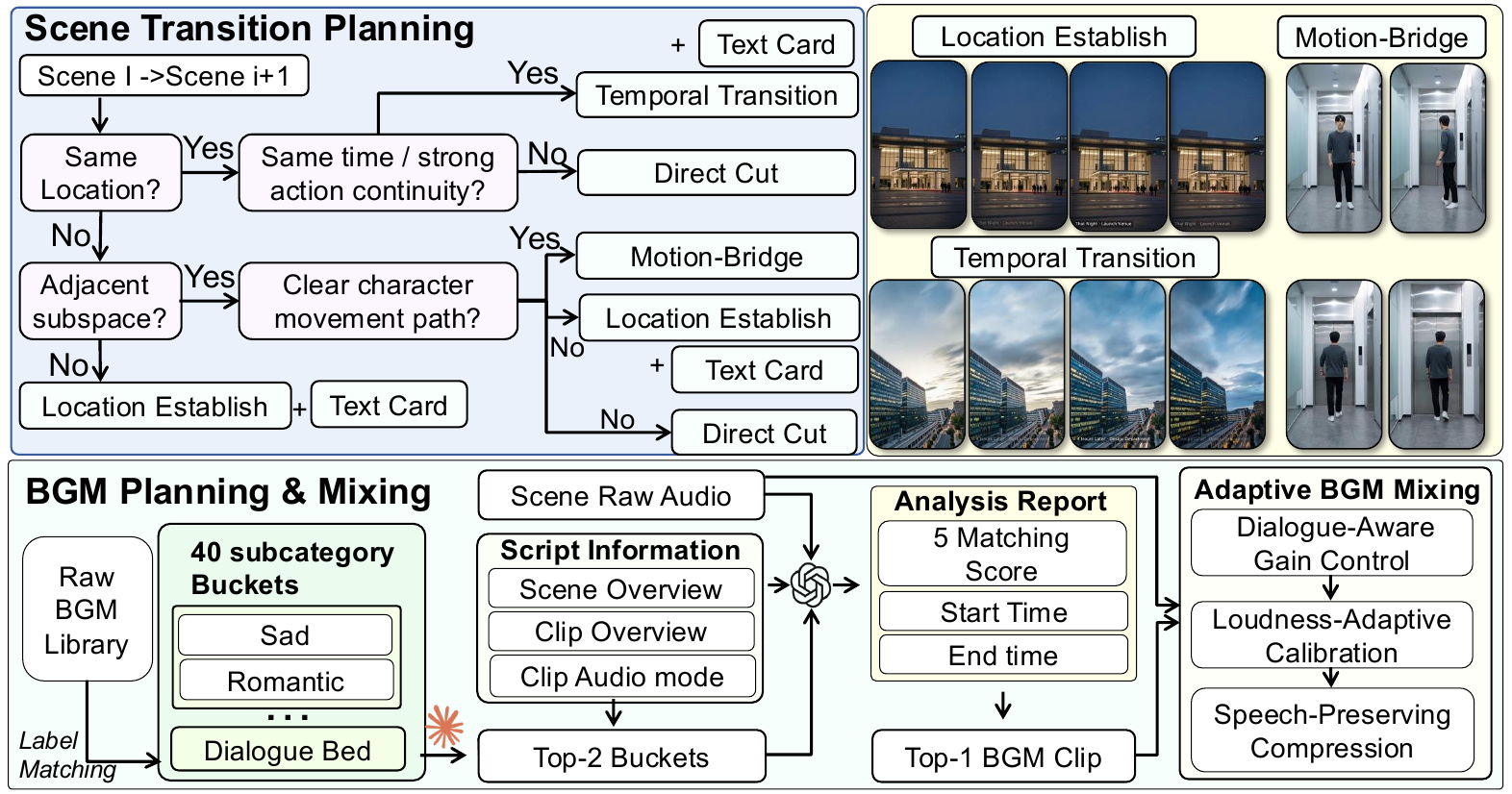}
\vspace{-4mm}
\caption{Our Diverse Transition Clips and BGM Planning \& Mixing}

\label{bgm_transition}
\vspace{-4mm}
\end{figure}

Built-in audio from video generation models often contains artifacts, mismatched music, or inconsistent background sound across clips. Since each scene in our pipeline consists of multiple generated clips, we add a scene-level BGM track to improve emotional consistency and reduce perceptual discontinuities. As shown in \cref{bgm_transition}, we first construct a short-drama-oriented BGM library organized into $8$ functional categories and $40$ fine-grained subcategories. Candidate tracks are assigned to these subcategories using provider-side metadata, including genre, vartag, instrument, and speed. For each scene, we use the scene overview, clip descriptions, clip-level BGM moods, and subcategory bucket descriptions to let an LLM select the most suitable primary and backup BGM buckets. We then call GPT-Audio to evaluate full candidate tracks from the selected subcategories buckets. Given the scene’s original audio and each candidate BGM, GPT-Audio predicts a scene-length segment and scores it by emotional fit, narrative fit, rhythm fit, and transition fit. The highest-scoring track segment is selected as the BGM for the entire scene.

Finally, we mix the selected BGM with the generated scene audio using adaptive volume control. We first lower the BGM base volume for dialogue-dense scenes, then calibrate the BGM level using the LUFS gap between the scene audio and BGM segment. We further apply speech-preserving dynamic compression so that BGM is reduced during dialogue-heavy regions and remains stronger in non-dialogue regions. This produces coherent scene-level music while preserving dialogue clarity.

\label[appendix]{app:transition_bgm}

\section{Multi-Character 3D-Consistent First-Frame Generation}
\label[appendix]{app:multi_character}
%倒着抽帧，直到人脸识别模型例如yolo或者sam啥的，检测到B，然后B同样进行一样的操作得到mask，3d人物，3d人物放入环境中，然后最后尾帧人物也放入，A继续走之前的流程，挑选出来Top-k（挑选仍然是老流程），然后来检测一下当前视角里面应不应该有B存在，有的，B也插入，没有就算了，然后A，B来贴图，加入人物颜色信息，后面的依旧？来选top-1
% 大部分仍然一样
% 只是在有多个人物存在的时候，并且在下一个clip，仍然需要这些人物存在的时候，为了更好地估计各个人物在环境中的位置，我们需要得到各个人物的大致位置，我们将会倒着抽帧，直到可以很好的取得对应的人物，继续通过sam 3d body提取人物3d模型，以及sam3 来提取对应的人物的位置，然后利用cut3r来输入从视频开始到此帧的视频，得到此帧的时候相机所在位置，从而吧人物3d模型注册到3d环境中，并且我们在这里会沿着相机和人物的线，来前后移动人物，要求再环境中漏出来的人物3d区域尽可能的和我们的帧提取出来的人物mask一致，从而更精确的把人物3d放入，对于尾帧人物，我们仍然是之前的步骤，来注册人物3d到世界3d模型中，这样，我们就有了多个人物3d模型在环境里面的位置

%另一个不同之处是，当我们在采取下一个clip的首帧候选的时候，我们会以多个人物的中心点，作为球心，来以不同的半径，不同的俯仰角，来进行采取候选帧，从而可以尽量保持多个人物都在画面里面，在我们后面的本地筛选中，我们会直接排除如果有人物的3d模型不在画面里的视角，这样我们可以尽量保证，主要人物在连续的2个clip中在环境里面的位置不会太大的变化
The main 3D-grounded first-frame generation pipeline in \cref{sec: keyframe to video generation} describes the single-character case. 
For multi-character clips, most steps remain unchanged: we still use the scene-level 3D world, video trajectory anchoring, geometry-aware camera sampling, character-conditioned first-frame generation, and frame-level review. The main difference is that we must first place all required characters into the same 3D world before sampling the next first-frame camera.

\textbf{Multi-Character Registration.}
Given two adjacent clips, we identify the set of characters that appear in the current clip and are still required in the next clip. 
The tail frame usually contains the primary character, which is registered to the 3D world using the procedure described in the main text. 
For other characters that are not clearly visible in the tail frame, we scan the current clip backward until finding a frame where the character is sufficiently visible and separable from the background. 
For each selected frame, SAM 3D Body~\cite{sam3d_body} reconstructs a character mesh and body keypoints, while SAM3~\cite{sam3} provides the corresponding person mask.

To localize this character in the scene-level 3D world, we recover the camera pose of the selected frame using CUT3R~\cite{wang2025continuous} on the video prefix ending at that frame, and anchor it to the same world coordinate system as in the single-character case. We then initialize the character transform from the reconstructed body keypoints and refine its depth along the camera--character ray. Specifically, we translate the mesh forward and backward along this ray and render it from the selected-frame camera. 
The final position is chosen such that the visible rendered silhouette best matches the SAM3 person mask, while preserving the 2D keypoint alignment. 
This refinement reduces depth ambiguity and gives a more reliable estimate of each character's position in the shared 3D scene. After this step, all involved characters are represented as 3D meshes registered in the same world coordinate system.

\textbf{Multi-Character Camera Sampling.}
Once all required characters are placed in the 3D world, we modify the next-shot camera sampling strategy. Instead of centering the spherical sampling region around a single character, we compute the center of all involved characters and use it as the camera target. Candidate cameras are then sampled with different radii, azimuths, and elevations around this multi-character center. This encourages the next first frame to keep the required characters inside the same view while preserving their relative spatial arrangement.

For each candidate camera, we render the scene background and all registered character meshes. The local geometric filter rejects views where any required character falls outside the image, becomes too small, is severely occluded, or has insufficient visible body/face area. It also removes cameras that are too close to scene surfaces or contain too little valid background. 
The remaining candidates are passed to the semantic VLM filter, which checks whether the rendered view supports the next-clip prompt, including scene anchors, character interaction direction, and relative blocking. 
The top-ranked views are then used for character-conditioned first-frame synthesis.

\textbf{First-Frame Synthesis and Review.}
For each selected view, the rendered multi-character meshes provide pose, scale, and spatial-layout constraints, while multi-view character references preserve identity and clothing. The image generation model synthesizes the next first frame conditioned on the rendered scene, the character references, and the previous clip context. Finally, the frame reviewer checks whether all required characters appear with correct identities, whether their relative positions remain consistent with the previous clip, and whether the background agrees with the scene-level 3D world. Frames that fail these checks are repaired or resampled. This extension allows the system to handle multi-character interactions while maintaining spatial continuity across adjacent clips.
\section{Human Rating Details}
We conduct a human rating study to complement model-based evaluation. 
We recruit $20$ annotators.
All participants receive monetary compensation at a rate consistent with local wage standards.
Before rating, participants are given written instructions describing the evaluation criteria, the 5-point Likert scale, and the meaning of each score, where 1 indicates very poor quality and $5$ indicates excellent quality with respect to the target criterion. 
All evaluated samples are anonymized: method names are removed, and participants are not informed which system generated each sample. 
For each metric, we use the same evaluation units as in the model-based protocol, 
including the opening segment of the drama for opening hook, scene-ending segments for end hook, scene-level segments for escalation effect, the complete drama for narrative coherence, adjacent clip boundaries for spatial and layout continuity, scene-level audio for BGM Emotion Alignment, and scene-boundary segments for transition naturalness. 
Samples from different methods are randomly shuffled for each participant to reduce ordering bias and method-specific bias. 
Each participant rates the assigned samples independently according to the corresponding short-drama criterion. 
For aggregation, we first average each participant's scores over all evaluation units belonging to the same method and metric, such as multiple clips, scene boundaries, or scene-level segments. 
We then average these per-participant scores across the $20$ participants to obtain the final human rating for each method and metric. 
The final human rating results are reported in \cref{tab:quantitative_and_human}.
\label[appendix]{app:Human Rating}

\section{Detailed Experiment Settings}
\label[appendix]{Baseline Setting}
%细节用的gpu，cpu资源，baseline的信息
All experiments are conducted with API-based image, video, language, and audio generation modules, together with local 3D and vision inference modules. 
For our method, the local components can be run on a single NVIDIA RTX A6000 GPU with $48$GB memory. The main GPU memory requirement comes from CUT3R, which is used for video trajectory estimation and frame pose anchoring. 
When a $48$GB GPU is unavailable, these local 3D modules can also be executed on CPU, but with substantially slower runtime. VGGT, SAM 3D Body, SAM3, and rendering are also executed locally. 

For baseline evaluation, most baselines can be run on the same $48$GB A6000 setup. 
LongLive, HoloCine, and StoryMem are run on an H200 GPU due to their higher memory and runtime requirements. All baselines are evaluated on the same $50$ prompts from Short-Drama-Bench using the evaluation protocol described in the main paper.

We use the same retry policy across generation stages. For text review, first-frame review, and generated video review, each failed item can be revised or regenerated at most three times. If the sample still fails after the maximum retry count, we keep the best available candidate according to the corresponding reviewer score. 
For 3D-consistent first-frame generation, we keep the top-$8$ candidate camera views after geometric and semantic filtering, and select the final first frame from these candidates using the reviewer score described in \cref{app:review_metrics}.

\textbf{Baseline selection and adoption.}
We compare against the most widely adopted public short-drama and story-visualization systems. 
ToonFlow is the most popular open-source short-drama workflow with 15.3k GitHub stars, followed by StoryMem (767), ScriptAgent (357), MovieAgent (358), LongLive (2.6k), and HoloCine (703); Xiao Yun Que is a leading commercial short-drama product with approximately 1.22M monthly visits.
GitHub statistics are as of September 2026, and monthly visits are as of December 2025.

ToonFlow~\cite{toonflow}: official release v1.1.6, default configuration.
Xiao Yun Que~\cite{xiaoyunque}: official public interface with the recommended configuration, accessed in March 2026.
LongLive~\cite{longlive}, HoloCine~\cite{holocine}, and StoryMem~\cite{storymem}: Claude Opus 4.6~\cite{anthropic2026claudeopus46} converts the single sentence into its required structured input, giving them more information than our method receives.
MovieAgent~\cite{movieagent} and ScriptAgent~\cite{scriptagent}: internal expansion LLMs replaced by Claude Opus 4.6.

\section{Time and Cost Analysis}
\textbf{Time Analysis.} We report approximate wall-clock generation time for producing one complete $10$ min short drama under our evaluation setting. 
The runtime depends on API latency, queueing time, video duration, and the number of reviewer-triggered retries, so the numbers should be interpreted as practical estimates rather than fixed constants.

For our method, story generation and multi-agent script refinement take about $10$--$15$ minutes. 
Image generation is performed through external APIs and can be parallelized across scenes and clips, so the first-frame, panorama, and visual-asset generation stage takes about $2$--$4$ minutes in practice. 
Scene-level 3D world construction takes about $2$--$4$ minutes per world. 
The dominant cost in wall-clock time is video generation, which takes nearly one hour for a typical short drama under our API setting. 
Overall, our pipeline usually takes about $74$--$90$ minutes to produce a complete short drama.

We also compare the practical runtime of representative baselines under the same benchmark prompts. 
Xiao Yun Que~\cite{xiaoyunque} typically takes about $1.5$--$2$ hours per drama, while Toonflow~\cite{toonflow} takes about $2$--$3$ hours. 
ScriptAgent~\cite{scriptagent} relies on API-based video generation but has a longer sequential pipeline, taking about $4$ hours. HoloCine~\cite{holocine} requires about 9 hours per drama. LongLive takes about 1 hour per drama.
MovieAgent~\cite{movieagent} and StoryMem~\cite{storymem} require heavier local generation and take about $35$ hours in our evaluation setting. 
These comparisons show that our framework improves generation quality while maintaining practical runtime, mainly because image and scene-level asset generation can be parallelized and local 3D inference is only a small fraction of the full pipeline time.

\textbf{Cost Analysis.} We estimate the API cost for generating a one-minute short drama under the $1080$P setting. 
For our method, the main API costs come from video generation, image generation, text/model review, and 3D world construction. 
Using Kling v3 Pro image-to-video at $\$0.168$/s, $60$ seconds of video generation costs $\$10.08$. 
Image generation uses about $30$ generated images, including first frames, panoramas, backgrounds, and repair images, costing about $\$4.02$ at $\$0.134$ per image. 
Text generation and review cost about $\$2.0$ per minute, and one World Labs Marble world costs about $\$1.2$. 
Without reviewer-triggered regeneration, the base cost is therefore about $\$17.3$/min. 
In practice, around half of the text, image, and video generation budget is spent on reviewer-triggered retries or repair, leading to an average cost of about $\$25$--$\$27$/min.

Compared with existing short-drama platforms, this cost is slightly higher but remains in a similar range. 
In our evaluation setting, Xiao Yun Que costs about $\$24.36$/min, while Toonflow costs about $\$21.53$/min, including approximately $\$15.51$ for video generation, $\$4.02$ for image generation, and $\$2.0$ for text generation. 
The additional cost of our method mainly comes from multi-stage review, regeneration after failed review, and 3D world construction. 
These costs improve quality and cross-clip consistency, but they remain an important target for future optimization.

The cost remains much lower than professional short-drama production. 
The public Dramaland quotation for Hongguo short dramas corresponds to about $\$293$/min for A-level productions, $\$439$/min for S-level productions, and $\$732$/min for S+ productions~\cite{dramaland_pricing}. 
For live-action productions with human actors, industry reports indicate a typical production cost of about $\$1{,}464$/min~\cite{sina_shortdrama_cost}. 
In comparison, our estimated API cost of about $\$25$--$\$27$/min is higher than existing automated short-drama platforms, but remains substantially lower than professional short-drama production while providing stronger controllability, reviewer-based refinement, and cross-clip spatial consistency.

\label[appendix]{Cost Analysis}

\section{Limitations}

\label[appendix]{Limitation}

Our framework has several limitations. 
First, the improved controllability and production quality come with higher generation cost. Because our system includes multi-stage generation, 3D world construction, automatic review, and retry mechanisms, its estimated API cost is higher than some existing short-drama platforms. For a one-minute generated drama, our average API cost is about $\$25$-$\$27$ per minute, while Xiao Yun Que costs about $\$24.36$ per minute under our evaluation setting. Although our system achieves better generation quality and Xiao Yun Que is a closed-source commercial platform whose internal pipeline and true operating cost are not fully observable, reducing cost remains important for large-scale deployment.

Second, the current framework emphasizes automatic generation and has limited human-in-the-loop interaction. Future systems could expose reviewer scores and diagnostic feedback to users through an interactive production interface. 
For example, clips with low reviewer scores could be automatically regenerated, clips with high scores could be accepted directly, and borderline cases could be routed to human creators for selecting whether and how to revise them. Such a hybrid workflow may reduce unnecessary retries while preserving production-level quality control.

Third, audio licensing is an important practical constraint for short-drama production. To reduce copyright risk, our current BGM library mainly contains royalty-free or commercially usable music, which limits the diversity of available styles and emotional expressions. A future system could integrate a larger licensed music library and provide users with explicit purchase or licensing options when a matched track is selected, improving audio quality while satisfying commercial publishing requirements.

Finally, 3D grounding inherits errors in estimated depth and pose when appearance or visible evidence changes. Synthetic-to-real robust depth estimation~\cite{yan2025synthetic} and weather-generalized adaptation~\cite{yan2025er} study related appearance-shift challenges in real imagery. Handling absent or unreliable cues is another relevant issue: FuseAdapt~\cite{zhang2026fuseadapt} addresses missing modalities in semantic segmentation. Adapting such robustness mechanisms to generated scenes and our geometry-dependent pipeline remains future work.

\section{Multi-Stage Review Metrics and Judge Models}
\label[appendix]{app:review_metrics}

Our system uses reviewer models in two settings. 
Internal reviewers are used during generation to trigger rewriting, regeneration, or candidate selection. 
External judges are used only after generation for benchmark evaluation. 
We summarize the metrics, judge models, and decision rules below.

\subsection{Text-Level Review}

\begin{table}[h]
\centering
\caption{Text-level reviewer configurations.}
\label{tab:text_review_metrics}
\setlength{\tabcolsep}{4pt}
\renewcommand{\arraystretch}{1.12}
\begin{tabular}{L{0.23\linewidth} L{0.70\linewidth}}
\toprule
\textbf{Stage} & \textbf{Configurations} \\
\midrule
Story bible / planning review
& \textbf{Role:} Evaluates the story core before clip expansion, focusing on narrative logic, short-drama pacing, hook design, and visual executability. \\
& \textbf{Judge models:} gpt-5.4-2026-03-05, claude-sonnet-4-6, gemini-3.1-pro-preview; Final Decider: gpt-5.4-pro-2026-03-05. \\
& \textbf{Metrics:} logic\_integrity, open\_grab, hook\_continuity, clarity, reversal\_pacing, payoff\_resolution, each scored $0$--$10$. \\
& \textbf{Gate:} visual\_executability\_gate in \{pass, borderline, fail\}. \\
& \textbf{Rule:} Low key scores, strong judge disagreement, or visual failure triggers arbiter-guided revision. Key thresholds: logic, open\_grab, and clarity $\geq 7$. \\
\midrule
Clip spatial text audit
& \textbf{Role:} Checks whether the clip-level text prompt is spatially executable before image or video generation. \\
& \textbf{Judge model:} claude-opus-4-6. \\
& \textbf{Metrics:} spatial\_consistency, physics\_plausibility, cross\_clip\_continuity, overall, each scored $0$--$10$. \\
& \textbf{Issue types:} POSITION, GAZE, ENTRY\_EXIT, PROP, CONTINUITY, OVERCROWDING. \\
& \textbf{Rule:} If pass=false, the system rewrites the corresponding prompt, state, or prop description. \\
\midrule
Prop continuity audit
& \textbf{Role:} Verifies whether key props have valid sources, movements, ownership changes, and end states across adjacent clips. \\
& \textbf{Judge model:} claude-opus-4-6. \\
& \textbf{Metrics:} prop\_source\_continuity, prop\_motion\_plausibility, overall, each scored $0$--$10$. \\
& \textbf{Rule:} If pass=false, the system applies a minimal prompt rewrite before generation. \\
\midrule
Next-clip scene-information audit
& \textbf{Role:} Determines from the next clip's textual description whether the next first frame requires additional scene information beyond the previous tail frame. \\
& \textbf{Judge model:} claude-opus-4-6. \\
& \textbf{Outputs:} needs\_extra\_scene\_information, has\_large\_character\_or\_camera\_reposition, required\_scene\_anchors. \\
& \textbf{Rule:} If needs\_extra\_scene\_information=true, the system does not simply reuse the previous tail frame and instead invokes the 3D-consistent first-frame generation path. \\
\bottomrule
\end{tabular}
\end{table}

\cref{tab:text_review_metrics} summarizes the text-level reviewers used before visual generation. 
These reviewers check story quality, prompt executability, prop continuity, and whether the next clip requires additional scene information for 3D-consistent first-frame generation.

\subsection{First-Frame and Tail-Frame Image Review}
\cref{tab:first_frame_review_metrics} summarizes the image-level reviewers used for first-frame selection and tail-frame routing. 
These reviewers select 3D-consistent first-frame candidates and determine whether a previous tail frame contains sufficient visual context for direct reuse.
\begin{table}[h]
\centering
\caption{First-frame and tail-frame image reviewer configurations.}
\label{tab:first_frame_review_metrics}
\setlength{\tabcolsep}{4pt}
\renewcommand{\arraystretch}{1.12}
\begin{tabular}{L{0.23\linewidth} L{0.70\linewidth}}
\toprule
\textbf{Stage} & \textbf{Configurations} \\
\midrule
3D-consistent first-frame candidate selection
& \textbf{Role:} Selects the best image candidate among 3D-consistent rendered views based on image quality, character integrity, and spatial continuity. \\
& \textbf{Judge model:} Qwen/Qwen3.5-397B-A17B. \\
& \textbf{Metrics:} temporal\_continuity, layout\_consistency, background\_quality, person\_scene\_interaction, character\_integrity, color\_continuity. \\
& \textbf{Score range:} each metric is scored $0$--$5$; total score is $0$--$30$. \\
& \textbf{Rule:} Any metric below $3$ rejects the candidate; among accepted candidates, the highest total score is selected. \\
\midrule
Tail-frame close-up detector
& \textbf{Role:} Checks whether the previous clip's tail frame is too local or context-poor to directly serve as the next first frame. \\
& \textbf{Judge model:} Qwen/Qwen3.5-397B-A17B. \\
& \textbf{Outputs:} is\_local\_closeup, shot\_scale, visible\_environment, confidence in $0$--$1$. \\
& \textbf{Rule:} If the tail frame is a local close-up or contains insufficient visible environment, it is not directly reused as the next first frame. \\

\bottomrule
\end{tabular}
\end{table}

\subsection{Video-Level Review}
\cref{tab:video_review_metrics} summarizes the video-level reviewers used after clip generation. 
They evaluate visual physics, temporal continuity, reaction plausibility, and whether character entrances, exits, and presence states match the clip script.
\begin{table}[h]
\centering
\caption{Generated video reviewer configurations.}
\label{tab:video_review_metrics}
\setlength{\tabcolsep}{4pt}
\renewcommand{\arraystretch}{1.12}
\begin{tabular}{L{0.23\linewidth} L{0.70\linewidth}}
\toprule
\textbf{Stage} & \textbf{Configurations} \\
\midrule
Generated video spatial audit
& \textbf{Role:} Checks whether the generated video is physically plausible and visually continuous across time, using sampled video frames rather than audio. \\
& \textbf{Judge model:} Qwen/Qwen3.5-397B-A17B, used as a visual/video reviewer over sampled frames. \\
& \textbf{Metrics:} physics\_integrity, temporal\_continuity, reaction\_plausibility, each scored 0--10. \\
& \textbf{Overall:} local overall is the mean of the above three metrics. \\
& \textbf{Rule:} Pass iff overall $\geq 5$ and character presence consistency passes; otherwise the system rewrites the prompt or regenerates the video. \\
\midrule
Character presence consistency
& \textbf{Role:} Verifies whether character presence, entrance, exit, and on-screen status follow the clip script. This metric does not evaluate identity similarity or action quality. \\
& \textbf{Judge model:} Qwen/Qwen3.5-397B-A17B, used as a visual/video reviewer over sampled frames. \\
& \textbf{Inputs:} characters\_at\_start, characters\_at\_end, and ordered character\_events from the clip script. \\
& \textbf{Checks:} whether the characters visible at the video beginning match characters\_at\_start; whether the characters visible at the video ending match characters\_at\_end; whether scripted entrances and exits occur in the correct order; whether any character appears without narrative support; whether a character remains visible after a scripted exit; and whether an unintended character dominates the frame. \\
& \textbf{Metric:} character\_presence\_consistency, binary score 0 or 10. \\
& \textbf{Rule:} The score must be 10; otherwise the video fails and is sent to prompt rewriting or regeneration. \\
\bottomrule
\end{tabular}
\end{table}

\subsection{Audio and BGM Review}
\cref{tab:audio_review_metrics} summarizes the audio reviewers used for scene-level BGM planning and selection. 
The system first selects suitable BGM subcategory buckets from textual scene context, then scores candidate audio segments to choose the final soundtrack.
\begin{table}[h]
\centering
\caption{BGM selection and audio reviewer configurations.}
\label{tab:audio_review_metrics}
\setlength{\tabcolsep}{4pt}
\renewcommand{\arraystretch}{1.12}
\begin{tabular}{L{0.23\linewidth} L{0.70\linewidth}}
\toprule
\textbf{Stage} & \textbf{Configurations} \\
\midrule
BGM bucket selection
& \textbf{Role:} Selects the most suitable functional BGM categories for each scene according to its narrative content and emotional trajectory. \\
& \textbf{Judge model:} claude-opus-4-6. \\
& \textbf{Inputs:} scene overview, clip descriptions, clip-level BGM moods, and subcategory bucket descriptions. \\
& \textbf{Outputs:} primary and backup BGM subcategory buckets. \\
& \textbf{Rule:} Candidate tracks are retrieved from the selected subcategory buckets before audio-level scoring. \\
\midrule
BGM segment scoring
& \textbf{Role:} Scores candidate BGM segments and selects the best segment/window for the scene-level soundtrack. \\
& \textbf{Judge model:} gpt-audio-2025-08-28. \\
& \textbf{Metrics:} emotion\_fit, narrative\_fit, rhythm\_fit, transition\_fit, each scored 0--10. \\
& \textbf{Rule:} the highest-scoring segment is selected. \\
\bottomrule
\end{tabular}
\end{table}

\subsection{External Benchmark Evaluation}
\cref{tab:external_eval_metrics} summarizes the judge models and metrics used only for final evaluation. 
These benchmark judges are separate from the internal reviewer loops and are applied to the generated videos from the same 50 Short-Drama-Bench prompts.
\begin{table}[h]
\centering
\caption{External benchmark evaluation settings.}
\label{tab:external_eval_metrics}
\setlength{\tabcolsep}{4pt}
\renewcommand{\arraystretch}{1.12}
\begin{tabular}{L{0.23\linewidth} L{0.70\linewidth}}
\toprule
\textbf{Benchmark} & \textbf{Configurations} \\
\midrule
Short-Drama-Bench
& \textbf{Role:} Evaluates short-drama-specific quality on the 50 benchmark prompts and their generated videos. \\
& \textbf{Judge models:} gemini-3.1-pro-preview, qwen3.5-omni-plus-2026-03-15, seed-2-0-pro-260328. \\
& \textbf{Metrics:} opening\_hook, end\_hook, escalation\_effect, narrative\_coherence, character\_spatial\_continuity, environment\_layout\_continuity, bgm\_emotion\_alignment, transition\_naturalness. \\
& \textbf{Score range:} each metric is scored 1--5. \\
& \textbf{Aggregation:} scores are averaged across judges, prompts, and evaluation units. \\
\midrule
VBench
& \textbf{Role:} Evaluates general video quality on the generated videos for the same 50 benchmark prompts. \\
& \textbf{Evaluator:} VBench automatic metric modules. \\
& \textbf{Metrics:} subject\_consistency, background\_consistency, motion\_smoothness. \\
& \textbf{Score range:} scores follow the native VBench normalization. \\
\midrule
ViStoryBench adaptation
& \textbf{Role:} Evaluates story-visualization quality on the generated videos for the same 50 benchmark prompts. \\
& \textbf{Judge model:} gpt-5.4-2026-03-05. \\
& \textbf{Metrics:} scene alignment, cross-character consistency, global character action, and single-character action. \\
& \textbf{Protocol:} we sample video frames at 1 FPS and apply the image-based ViStoryBench story-visualization criteria to the sampled frames. \\
& \textbf{Aggregation:} scores are averaged across prompts, scenes, and sampled frames. \\
\bottomrule
\end{tabular}
\end{table}

\clearpage
\newpage

\section{Reliability and Sensitivity Analysis}
\label[appendix]{app:reliability_sensitivity}

This section evaluates the reliability of our evaluation protocol and
examines whether the main conclusions are sensitive to evaluator choice. We analyze four aspects: agreement between
human and LLM-based evaluation, robustness to the choice of LLM judge,
confidence intervals for the improvement over the strongest baseline,
and uncertainty and inter-rater reliability in the human study.

\subsection{Human--LLM Evaluation Agreement}
\label{subsec:human_llm_agreement}

We first examine whether the LLM-based evaluation reflects human judgments. 
% We pair the human ratings and LLM-judge scores for the same
% eight methods and eight Short-Drama-Bench metrics. For each individual
% metric, the correlation is computed over the six method-level average
% scores. The pooled correlation is computed over all
% $8 \times 8 = 64$ method--metric pairs.
For each of the 50 prompts and each of the eight Short-Drama-Bench metrics, we average the ratings from 20 raters and the reference scores from three LLM judges separately for each of the eight methods.
We then compute Pearson and Spearman correlations across the eight paired method scores within each prompt. 

As shown in  \cref{tab:human_llm_agreement}, the median prompt-level Pearson correlations range from 0.878 to 0.960, while the median Spearman correlations range from 0.848 to 0.949 across the eight metrics.
Across all 3,200 method–prompt–metric pairs, the LLM-judge reference scores are approximately 0.38 points higher than the human ratings on average.

% As shown in \cref{tab:human_llm_agreement}, the pooled Pearson
% correlation is $r=0.982$, with a 95\% bootstrap confidence interval of
% $[0.965, 0.995]$, while the pooled Spearman correlation is
% $\rho=0.982$. At the pooled level, Kendall's correlation is
% $\tau=0.927$. Agreement is consistently high across the eight
% metrics, although it is lower for BGM Emotion Alignment, which involves
% a more subjective assessment of whether the music matches the intended
% scene emotion. We additionally observe a calibration difference: the
% LLM judges assign scores approximately $0.37$ points higher than the
% human evaluators on average.

\begin{table}[h]
\centering
\caption{
Prompt-level correlations between human ratings and LLM-judge reference scores across the eight Short-Drama-Bench metrics. For each prompt and metric, correlations are computed across eight methods.
}
\label{tab:human_llm_agreement}
\setlength{\tabcolsep}{7pt}
\renewcommand{\arraystretch}{1.12}
\begin{tabular}{L{0.55\linewidth}cc}
\toprule
\textbf{Evaluation metric}
& \textbf{Pearson $r$}
& \textbf{Spearman $\rho$} \\
\midrule
Opening Hook
& 0.960 & 0.929 \\

End Hook
& 0.909 & 0.892 \\

Escalation Effect
& 0.950 & 0.931 \\

Narrative Coherence
& 0.955 & 0.949 \\

Character Spatial Continuity
& 0.878 & 0.848 \\

Environment Layout Continuity
& 0.921 & 0.884 \\

BGM Emotion Alignment
& 0.936 & 0.881 \\

Transition Naturalness
& 0.958 & 0.883 \\
\bottomrule
\end{tabular}
\end{table}

\subsection{Robustness to LLM Judge Choice}
\label{subsec:judge_choice_robustness}

We next examine whether the benchmark conclusion depends on the choice of a particular LLM judge. Using the same generated dramas, we score all eight methods with three judge models. 
As shown in \cref{tab:judge_choice_sensitivity}, the judges assign different absolute scores but agree on the top three methods: Ours, Xiao Yun Que, ToonFlow.

Among the three judge models used in the main evaluation---Gemini, Qwen, and Seed---the median pairwise item-level Spearman correlation is $0.823$, with a range of $0.800$--$0.839$. Krippendorff's $\alpha$ among these three judges is $0.804$. These results indicate
that the ranking of the leading methods is robust to the choice of LLM judge, despite differences in score calibration.

\begin{table}[h]
\centering
\caption{
Average Short-Drama-Bench scores assigned by three LLM judges.
All three judges rank ours first, followed by Xiao Yun Que and ToonFlow.
}
\label{tab:judge_choice_sensitivity}
\small
\setlength{\tabcolsep}{4pt}
\renewcommand{\arraystretch}{1.10}
\resizebox{\linewidth}{!}{
\begin{tabular}{lcccccccc}
\toprule
\textbf{Judge model}
& \textbf{MovieAgent}
& \textbf{ScriptAgent}
& \textbf{StoryMem}
& \textbf{LongLive}
& \textbf{HoloCine}
& \textbf{ToonFlow}
& \textbf{Xiao Yun Que}
& \textbf{Ours} \\
\midrule
Gemini 3.1 Pro Preview
& 1.82 & 2.57 & 2.42  & 2.12  & 2.65 & 3.25 & 3.48 & \textbf{3.84} \\
Qwen3.5-Omni-Plus
& 2.10 & 2.85 & 2.73 & 1.92 & 2.44 & 3.56 & 3.78 & \textbf{4.18} \\
Seed 2.0 Pro
& 1.93 & 2.68 & 2.54 & 2.36  & 2.72 & 3.36 & 3.60 & \textbf{3.97} \\
\bottomrule
\end{tabular}
}
\end{table}

\subsection{Statistical Confidence of Performance Improvements}
\label{subsec:performance_confidence}

We quantify the statistical uncertainty of the improvement over the
strongest baseline using paired resampling. For the LLM-based
evaluation, we perform paired bootstrap resampling over the 50
Short-Drama-Bench prompts with $B=10{,}000$ bootstrap samples. For the
human evaluation, the confidence interval accounts for variation across
both prompts and evaluators. Xiao Yun Que is the strongest baseline in
both evaluation settings.

As reported in \cref{tab:performance_confidence_intervals}, our method
improves the average LLM-judge score by $0.375$, with a 95\% confidence
interval of $[0.282, 0.470]$. In the independent human evaluation, our
method improves the average score by $0.349$, with a 95\% confidence
interval of $[0.191, 0.515]$. Both confidence intervals remain entirely
above zero, showing that the improvements are robust to prompt and
evaluator variation.

\begin{table}[h]
\centering
\caption{
Score improvements of our method over Xiao Yun Que, the strongest
baseline in both LLM-based and human evaluations. The 95\% confidence
intervals are computed using paired resampling, and positive differences
favor our method.
}
\label{tab:performance_confidence_intervals}
\setlength{\tabcolsep}{8pt}
\renewcommand{\arraystretch}{1.12}
\begin{tabular}{L{0.38\linewidth}cc}
\toprule
\textbf{Evaluation setting}
& \textbf{Score difference}
& \textbf{95\% confidence interval} \\
\midrule
LLM-based evaluation
& \textbf{0.375}
& $[0.282,\,0.470]$ \\

Human evaluation
& \textbf{0.349}
& $[0.191,\,0.515]$ \\
\bottomrule
\end{tabular}
\end{table}

\subsection{Uncertainty and Inter-Rater Reliability in Human Evaluation}
\label{subsec:human_evaluation_reliability}

The human study includes 20 evaluators and 50 prompts, resulting in
1,000 ratings for each evaluated method. The scores in
\cref{tab:human_evaluation_statistics} are averaged over Opening Hook,
End Hook, Escalation Effect, Narrative Coherence, Character Spatial
Continuity, Environment Layout Continuity, BGM Emotion Alignment, and
Transition Naturalness. The confidence intervals account for variation
across both evaluators and prompts.

We additionally report ICC$(2,20)$, which measures the reliability of
the aggregated ratings from the 20 evaluators. The ICC$(2,20)$ values
range from $0.79$ to $0.86$, indicating consistent aggregated human
judgments. Our method achieves the highest human-evaluation mean,
obtaining $3.52$.

\begin{table}[h]
\centering
\caption{
Human-evaluation means, standard deviations, 95\% confidence intervals, and inter-rater reliability for the eight evaluated methods.
Each method receives 1,000 ratings from 20 evaluators over 50 prompts.
}
\label{tab:human_evaluation_statistics}
\setlength{\tabcolsep}{8pt}
\renewcommand{\arraystretch}{1.12}
\begin{tabular}{lcccc}
\toprule
\textbf{Method}
& \textbf{Mean}
& \textbf{SD}
& \textbf{95\% CI}
& \textbf{ICC$(2,20)$} \\
\midrule
MovieAgent
& 1.67 & 0.70 & $[1.53, 1.81]$ & 0.83 \\

ScriptAgent
& 2.28 & 0.77 & $[2.12, 2.45]$ & 0.79 \\

StoryMem
& 2.27 & 0.78 & $[2.11, 2.45]$ & 0.82 \\

LongLive
& 1.79 & 0.72 & $[1.68, 1.91]$ & 0.85 \\

HoloCine
& 2.24 & 0.82 & $[2.11, 2.39]$ & 0.86 \\

ToonFlow
& 2.97 & 0.85 & $[2.79, 3.16]$ & 0.82 \\

Xiao Yun Que
& 3.17 & 0.86 & $[2.99, 3.35]$ & 0.85 \\

\textbf{Ours}
& \textbf{3.52}
& 0.79
& $[3.35, 3.70]$
& 0.81 \\
\bottomrule
\end{tabular}
\end{table}

\section{Isolation Ablations of the Idea Bank and Geometry-Adaptive Registration}
\label[appendix]{app:isolation_ablation}

Here we isolate the two algorithm-level designs, keeping all other settings identical.

\textbf{Idea Bank with final-round revival.}
We generate scripts with and without the Idea Bank under identical settings and score them with the story-bible reviewer of \cref{tab:text_review_metrics} on its six dimensions ($0$--$10$).
As shown in \cref{tab:idea_bank_ablation}, the gains concentrate on the memorability-related dimensions (payoff $+0.2$, reversal pacing $+0.2$, opening
and hook continuity $+0.1$), matching the design goal of restoring strong elements without harming logic.
The over-correction it addresses is not specific to short dramas and applies to any LLM-based iterative narrative creation, such as story writing or plot planning for games.

\begin{table}[h]
\centering
\caption{Isolation ablation of the Idea Bank (story-bible reviewer scores, $0$--$10$).}
\label{tab:idea_bank_ablation}
\setlength{\tabcolsep}{5pt}
\begin{tabular}{l cccccc}
\toprule
\textbf{Setting} & \textbf{Logic} & \textbf{Opening} & \textbf{Hook cont.} & \textbf{Clarity} & \textbf{Reversal} & \textbf{Payoff} \\
\midrule
w/o Idea Bank & 8.17 & 8.98 & 8.87 & 8.32 & 8.59 & 8.61 \\
\textbf{w/ Idea Bank (Ours)} & \textbf{8.17} & \textbf{9.07} & \textbf{8.97} & \textbf{8.37} & \textbf{8.77} & \textbf{8.82} \\
\bottomrule
\end{tabular}
\end{table}

\textbf{Geometry-adaptive character registration.}
The ablation baseline projects the reconstructed mesh onto the registration frame and adjusts its placement by grid search so that the projected silhouette and keypoints match the person in the image; since enlarging the mesh while moving it proportionally farther away leaves the projection unchanged, this alignment cannot determine the character's distance. Our method shares every other step and additionally solves the distance from scene geometry.
We measure placement accuracy with three metrics: 
\emph{scene penetration rate}, the fraction of the $70$ body keypoints that fall inside scene solids such as the floor or tabletops; 
\emph{foot-to-floor distance}, the distance from the foot keypoints to the floor plane, reported as a fraction of the character's height, which grows whether the
character sinks or floats; and 
\emph{visible-region IoU}, the overlap between the visible part of the character rendered inside the 3D world and the person region visible in the source frame, which drops when incorrect placement changes how furniture occludes the character.
As shown in \cref{tab:registration_ablation}, our registration nearly eliminates penetration and floating and substantially improves occlusion consistency.
The same procedure applies to placing a single-image human reconstruction into any virtual scene, e.g., for world models or VR/AR content.

\begin{table}[h]
\centering
\caption{Isolation ablation of the geometry-adaptive registration. }
\label{tab:registration_ablation}
\setlength{\tabcolsep}{8pt}
\begin{tabular}{l cc}
\toprule
\textbf{Metric} & \textbf{Projection-only baseline} & \textbf{Ours} \\
\midrule
Scene penetration rate $\downarrow$ & 21.4\% & \textbf{0.7\%} \\
Foot-to-floor distance (\% of human height) $\downarrow$ & 63.2\% & \textbf{3.9\%} \\
Visible-region IoU $\uparrow$ & 0.27 & \textbf{0.67} \\
\bottomrule
\end{tabular}
\end{table}

\section{Short-Drama-Bench Prompts and Generated Videos}
\label[appendix]{app:benchmark_samples}

Table~\ref{tab:short-drama-topics} lists the prompt topics used in Short-Drama-Bench. 
The benchmark covers seven high-level short-drama genres, including underdog comeback, social realism, ancient court intrigue, suspense and thriller, time travel and rebirth, sweet romance, and corporate/business war. 
Each genre is further divided into fine-grained subcategories, with each subcategory containing representative story prompts. 
These prompts are designed to evaluate whether the generation pipeline can handle diverse narrative settings, character relationships, conflict structures, and genre-specific storytelling patterns.

\begin{longtable}{p{0.22\textwidth} p{0.22\textwidth} p{0.50\textwidth}}
\toprule
Category & Subcategory & Specific Topics \\
\midrule
\endfirsthead

\toprule
Category & Subcategory & Specific Topics \\
\midrule
\endhead

Underdog Comeback & Marriage Comeback & 1. My Unremarkable Husband Turns Out to Be the Company Chairman \\
 &  & 2. After Being Abandoned at the Wedding, She Returned as an Investor Capable of Buying the Groom's Empire \\
 &  & 3. The Day the Divorce Papers Were Signed, His Ex-Wife's Company Went Public \\
 &  & 4. The Daughter-in-Law Kicked Out by Her Mother-in-Law Became the New Owner of Her Company Three Years Later \\
\cmidrule(l){2-3}
 & Hidden Identity & 1. The Humiliated Stable Boy Turns Out to Be the Long-Lost Heir to the Kingdom \\
 &  & 2. The Security Guard Everyone in the Company Looks Down On Has Five World Leaders' Private Numbers in His Phone \\
 &  & 3. The Transfer Student Mocked by Classmates Whose Father Is Their School's Chairman of the Board \\
\cmidrule(l){2-3}
 & Career Comeback & 1. The designer's wife, who was accused of stealing her boss's manuscript, was confronted by her husband who had been secretly married. With a single phone call, the CEO was summoned. \\
 &  & 2. The Intern Publicly Humiliated by the Director Was Sitting in the Director's Chair a Year Later \\
 &  & 3. The Designer Reduced to Tears by a Client Went On to Win an International Design Award \\

\midrule
Social Realism & Workplace Injustice & 1. The Woman Fired for Being Pregnant Returned as the Company's Biggest Client \\
 &  & 2. The Middle Manager `Optimized Out' at 35 Built a Startup Team from an Unemployment Group Chat \\
 &  & 3. The Engineer Forced to Sign a Non-Compete Discovered the Boss Had Already Violated His Own \\
\cmidrule(l){2-3}
 & Medical \& Survival & 1. The Night the Hospital Refused Her Surgery, She Livestreamed Everything \\
 &  & 2. Her Father's Life-Saving Pill Costs 700 Yuan Each, So the Daughter Went to India to Find the Manufacturer Herself \\
 &  & 3. In the Three Months She Was Misdiagnosed with Cancer, She Saw Everyone Around Her for Who They Really Are \\
\cmidrule(l){2-3}
 & Family Ethics & 1. When the Mother Who Favored Sons Over Daughters Fell Ill, Only the Neglected Daughter Came \\
 &  & 2. The Parents Gave the House to Their Son but Left the Debt to Their Daughter \\
 &  & 3. The Whole Family Pooled Money for the Brother to Study Abroad, but the Sister Got Into a Better School on Her Own \\

\midrule
Ancient Court Intrigue & Harem Power Struggle & 1. The Abandoned Consort in the Cold Palace Is Determined to Put the Crown Prince on the Throne \\
 &  & 2. Sentenced to Death on Her First Day in the Palace, She Traded a Bowl of Poison for the Empress's Secret \\
 &  & 3. She Pretended to Be Out of Favor for Three Years While Secretly Building a Shadow Guard That Answers Only to Her \\
\cmidrule(l){2-3}
 & Court Conspiracy & 1. The Poisoned Princess Married the Enemy Prince Only to Burn the Empire from Within \\
 &  & 2. Everyone Believed the Chancellor Was Loyal --- Only the Crown Prince Knew He Killed the Late Emperor \\
 &  & 3. The Exiled General's Daughter Returns with Her Father's Former Army \\
\cmidrule(l){2-3}
 & Women Breaking the Rules & 1. She Disguised Herself as a Man to Top the Imperial Exam, Only to Be Exposed in the Golden Hall \\
 &  & 2. The Princess Who Knew No Martial Arts Talked Down a Hundred Thousand Rebels with Words Alone \\
 &  & 3. The Merchant's Daughter Who Married into the General's Household Brought Down Military Corruption with Her Ledger \\

\midrule
Suspense \& Thriller & Digital-Age Thriller & 1. The Serial Killer Had Been Lurking in Their Group Chat All Along \\
 &  & 2. The Missing Intern Sent a Video from the CEO's Private Basement \\
 &  & 3. The Social Media Post She Deleted Became the Only Lead in the Case \\
\cmidrule(l){2-3}
 & Closed-Space Mystery & 1. In a Snowed-In Mountain Lodge, One of the Eight Guests Is a Fugitive from Ten Years Ago \\
 &  & 2. An Elevator Malfunction Traps Six People --- One of Them Has a Bloody Knife in Their Bag \\
 &  & 3. Halfway Through a Murder Mystery Game, Someone Realizes the Script Is Based on a Real Person in the Room \\
\cmidrule(l){2-3}
 & Twist Thriller & 1. She Called the Police to Report Her Husband Missing, but They Found a Clue in Her Own Car Trunk \\
 &  & 2. Three Women Went Missing in a Row --- the Person Who Filed the Report Turned Out to Be the Suspect \\
 &  & 3. The Therapist Discovered That Her Most Dangerous Patient Is Actually Her Own Husband \\

\midrule
Time Travel \& Rebirth & Professional Time Travel & 1. A Modern Medical Student Travels Back to the Late Han Dynasty to Practice Medicine \\
 &  & 2. A Chemistry PhD Travels to an Era of Witches, Kings, and Poisons \\
 &  & 3. A Modern Forensic Scientist Travels to Ancient Times and Overturns a Wrongful Conviction Through Autopsy \\
\cmidrule(l){2-3}
 & Rebirth \& Revenge & 1. A Female Lawyer Is Reborn as a Queen Accused of Treason \\
 &  & 2. She Is Reborn to the Day Before Her Murder --- This Time She Gathers All the Evidence First \\
 &  & 3. After Rebirth She Didn't Rush to Seek Revenge --- She First Got Admitted to Law School \\

\midrule
Sweet Romance & Status Gap / Contract Romance & 1. She Fake-Married Her Roommate for a Visa --- Then It Stopped Being Fake \\
 &  & 2. The CEO's Blind Date Turns Out to Be the Girl He Anonymously Sponsored for Ten Years \\
\cmidrule(l){2-3}
 & Reconciliation Romance & 1. Five Years After Breaking Up, They Reunite in the ER --- She's His Attending Physician \\
 &  & 2. He Finally Found His First Love, but She No Longer Remembers Him \\

\midrule
Corporate \& Business War & Business Showdown & 1. On Her First Day at Work, She Discovered the Company's Biggest Corporate Spy Is Her Own Mentor \\
 &  & 2. Two Interns Both Fell for the Same Proposal --- One Chose to Plagiarize, the Other Chose to Innovate \\
 &  & 3. The Founder Kicked Out by His Partners Started Over --- Taking the Core Technology with Him \\

\bottomrule
\caption{Short-Drama-Bench prompt topics.}
\label{tab:short-drama-topics}
\end{longtable}

\begin{figure}[h]
\centering
\includegraphics[width=1.0\textwidth]{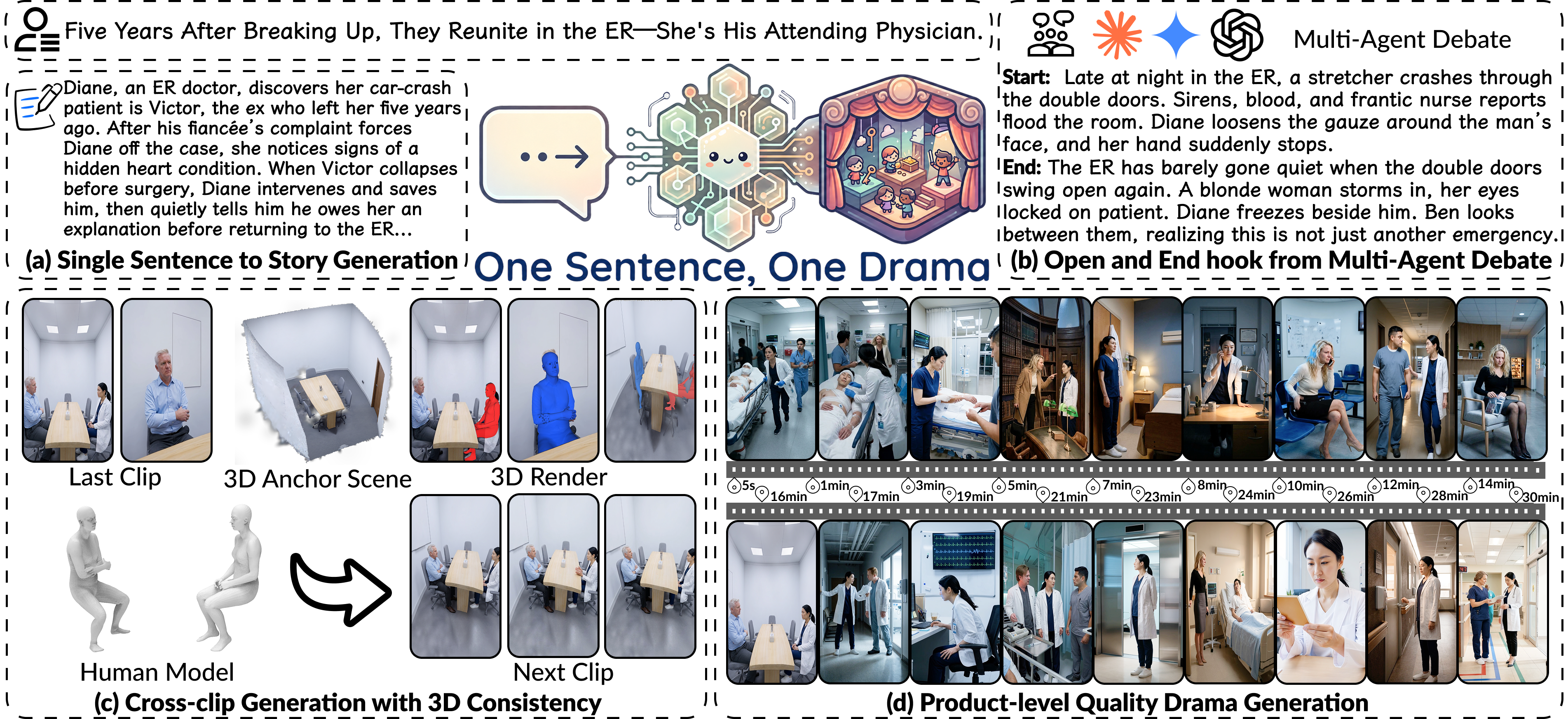}
\caption{Representative short dramas generated by our framework from single-sentence ideas, covering different genres, durations, and character settings.}
\label{fig:teaser}
\end{figure}

\begin{figure}[h]
\centering
\includegraphics[width=1.0\textwidth]{Figures/VideoGallery1_light.pdf}
\caption{Representative generated videos (Gallery 1/2).}
\label{VideoGallery1}
\end{figure}

\begin{figure}[h]
\centering
\includegraphics[width=1.0\textwidth]{Figures/VideoGallery2_light.pdf}
\caption{Representative generated videos (Gallery 2/2).}
\label{VideoGallery2}
\end{figure}

\cref{fig:teaser}, \cref{VideoGallery1} and \cref{VideoGallery2} present representative generated video examples from Short-Drama-Bench, illustrating the visual results produced under different prompt topics and narrative settings.

\section{Script Library and BGM Library Details}
\label[appendix]{app:bgm_bank}
To strengthen narrative planning, we use a short-drama database that includes $1{,}102$ short-drama scripts released under the MIT License, which are distilled into $16{,}311$ beat cards and $4{,}750$ logic chunks. This structured database provides retrieval-based references for plot rhythm, conflict escalation, and genre-specific storytelling patterns, giving the generation pipeline stronger narrative logic and short-drama style priors.

\begin{figure}[h]
\centering
\includegraphics[width=0.8\textwidth]{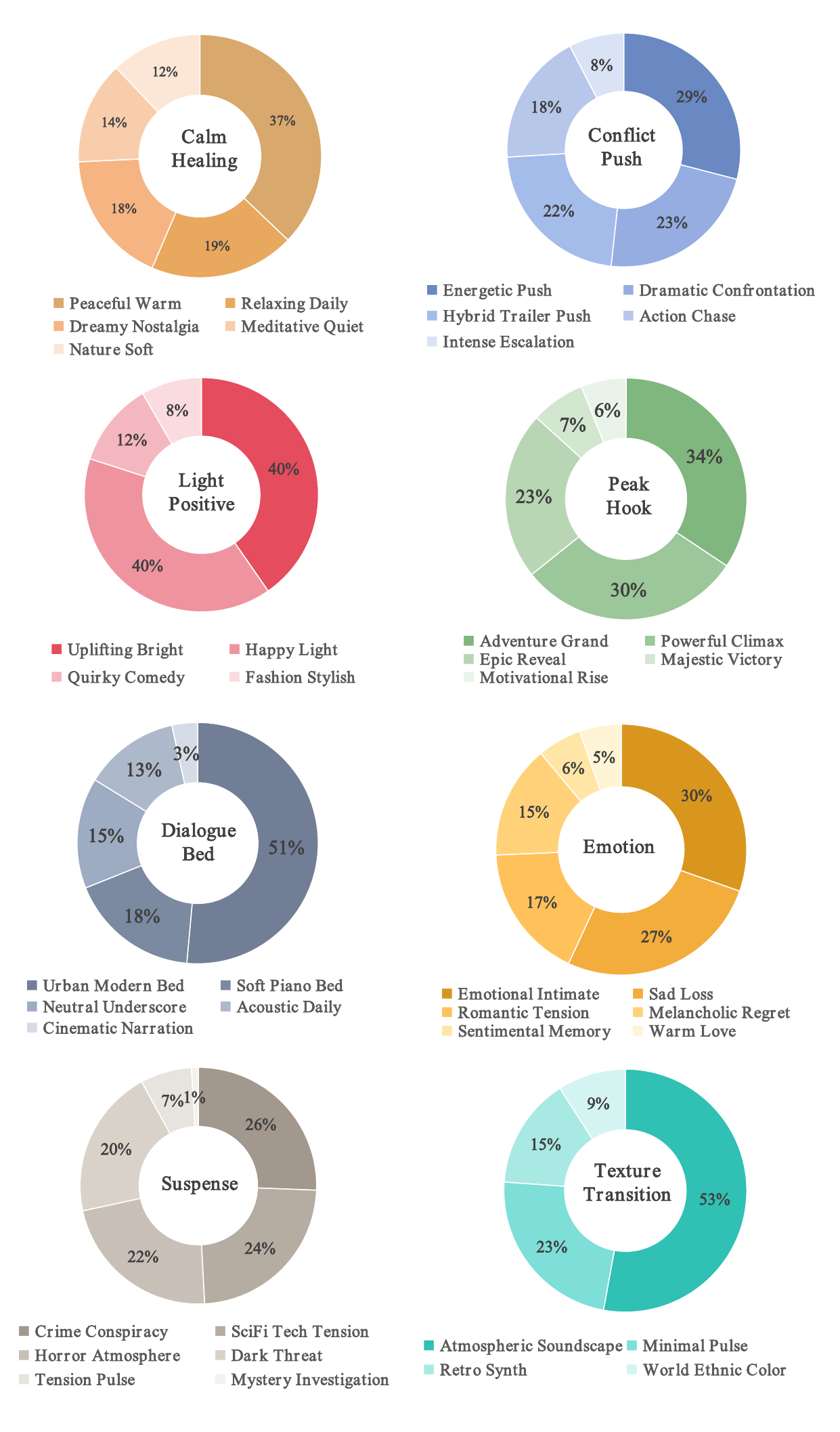}
\caption{Our BGM Datasets}
\label{bgm_data}
\end{figure}

As shown in \cref{bgm_data}, we build a BGM library with $8{,}122$ tracks, covering 8 high-level categories and $40$ fine-grained subcategories; each category is paired with a textual description that guides subsequent audio matching according to the scene rhythm and emotional intent.

\label[appendix]{app:bgm_datasets}

\section{Prompt Templates for Each Stage}

\label[appendix]{app:prompts}

\subsection{Evaluation Prompt}
\cref{fig:eval_prompt_template} shows the unified prompt template used for model-based Short-Drama-Bench evaluation. 
The same template is instantiated with different metrics, scopes, rubrics, and reference contexts to evaluate narrative quality, continuity, audio alignment, and transition naturalness.

\begin{figure}[!h]
\centering
\begin{tcolorbox}[
    width=0.98\linewidth,
    colback=gray!5,
    colframe=blue!45!black,
    colbacktitle=blue!18,
    coltitle=black,
    title={Short-Drama-Bench Evaluation Prompt Template},
    fonttitle=\bfseries\centering,
    arc=2mm,
    boxrule=0.8pt,
    left=2mm,
    right=2mm,
    top=2mm,
    bottom=2mm
]
\small
You are a rigorous short-drama video benchmark judge.

\vspace{1mm}
\textbf{Metric:} \texttt{\{metric\}} \\
\textbf{Scope:} \texttt{\{scope\}} \\
\textbf{Rubric:} \texttt{\{rubric\}}

\vspace{1mm}
Use a $1$--$5$ integer score: \\
$1$ = weak / mostly ineffective \\
$2$ = partially effective but with clear issues \\
$3$ = mostly effective with minor issues \\
$4$ = excellent and clearly satisfies the metric \\
$5$ = exceptional; reserve this only for outstanding cases with strong evidence, almost no visible weakness, and especially strong short-drama impact

\vspace{1mm}
\textbf{Reference context:} \\
\texttt{\{context\}}

\vspace{1mm}
\textbf{Return strict JSON only:} \\
\texttt{\{"score": $1$, "analysis": "brief evidence-based explanation"\}}

\vspace{2mm}
\textbf{Metric-specific fields:} \\
\texttt{opening\_hook} / drama: Evaluate whether the first $10$ seconds quickly attract viewers, establish stakes, and create immediate curiosity. \\
\texttt{end\_hook} / scene: Evaluate whether the scene ending creates sufficient motivation to continue watching. \\
\texttt{escalation\_effect} / scene: Evaluate whether the middle portion contains effective conflict escalation, rising pressure, or power reversal. \\
\texttt{narrative\_coherence} / drama: Evaluate whether the story logic remains clear, non-confusing, and internally consistent. \\
\texttt{character\_spatial\_continuity} / clip\_pair: Evaluate adjacent clip continuity, including character identity, screen position, body orientation, and spatial relationship continuity. \\
\texttt{environment\_layout\_continuity} / clip\_pair: Evaluate adjacent clip continuity, including background layout, room geometry, object placement, and camera-direction consistency. \\
\texttt{bgm\_emotion\_alignment} / scene: Evaluate whether the background music emotion aligns with scene mood, tension, pacing, and dramatic intent. \\
\texttt{transition\_naturalness} / scene\_pair: Evaluate whether the previous scene tail, transition clip, and next scene head connect naturally in action, camera, mood, and story logic.
\end{tcolorbox}
\caption{Prompt template for model-based Short-Drama-Bench evaluation.}
\label{fig:eval_prompt_template}
\end{figure}

\subsection{Text Review}

\begin{figure}[!h]
\centering
\begin{tcolorbox}[
    width=0.98\linewidth,
    colback=gray!5,
    colframe=blue!45!black,
    colbacktitle=blue!18,
    coltitle=black,
    title={Scene-Level Script Review Prompt Template},
    fonttitle=\bfseries\centering,
    arc=2mm,
    boxrule=0.8pt,
    left=2mm,
    right=2mm,
    top=2mm,
    bottom=2mm
]
\small
You are a reviewer for viral short-drama scripts designed for AI video production.

\vspace{1mm}
You will receive the full story information and all clips of a single scene in structured JSON format. \\
Your task is to output scores, reasons, and improvement suggestions for three metrics: \\
1) \texttt{hook}: evaluate only the opening attraction of the first clip. \\
2) \texttt{scene\_end}: evaluate only the ending hook of the last clip. \\
3) \texttt{twist}: evaluate only the reversal density of the middle clips, excluding the first and last clips.

\vspace{1mm}
\textbf{Boundary rules:} \\
- When scoring \texttt{hook}, do not use the last clip or \texttt{ending\_hook} as supporting evidence. \\
- When scoring \texttt{scene\_end}, do not use the opening attraction of the first clip as supporting evidence. \\
- When scoring \texttt{twist}, do not use the first or last clip as the main evidence. \\
- If the scene has fewer than $3$ clips, set \texttt{twist.score} to $8$ and \texttt{twist.improvements} to an empty array.

\vspace{1mm}
\textbf{Number of improvement suggestions:} \\
- Score $1$--$3$: exactly $3$ improvement suggestions. \\
- Score $4$--$6$: exactly $2$ improvement suggestions. \\
- Score $7$: exactly $1$ improvement suggestion. \\
- Score $8$--$10$: exactly $0$ improvement suggestions.

\vspace{1mm}
\textbf{Scoring rubric:} \\
\texttt{hook}: $1$--$3$ = flat opening, weak conflict, cannot attract viewers; $4$--$6$ = conflict exists but intensity or pacing is insufficient; $7$ = acceptable and can retain viewers; $8$--$10$ = highly attractive, with clear conflict and emotional momentum. \\
\texttt{scene\_end}: $1$--$3$ = no escalation or suspense; $4$--$6$ = ending intention exists but the hook is weak; $7$ = clear ending hook; $8$--$10$ = strong escalation and clear anticipation for the next scene. \\
\texttt{twist}: $1$--$3$ = little or no middle reversal, or mostly repetitive progression; $4$--$6$ = some turns exist but density or quality is moderate; $7$ = acceptable middle-clip turning points; $8$--$10$ = high reversal density and strong reversal quality.

\vspace{1mm}
\textbf{Additional constraints:} \\
- Review only; do not rewrite the script. \\
- Improvement suggestions must be actionable and specific. \\
- Avoid text-dependent visual solutions, wet-body/water-stain descriptions, and ellipses in dialogue. \\
- All strings must be single-line strings; if a line break is needed, use \texttt{\textbackslash n}. \\
- Return strict JSON only, without Markdown or explanation.

\vspace{1mm}
\textbf{Output format:}
\begin{verbatim}
{
  "hook": {"score": 1-10, "reason": "...", "improvements": ["..."]},
  "scene_end": {"score": 1-10, "reason": "...", "improvements": ["..."]},
  "twist": {"score": 1-10, "reason": "...", "improvements": ["..."]}
}
\end{verbatim}

\textbf{User prompt template:}
\begin{verbatim}
Please output the hook/scene_end/twist review JSON for this scene:
{payload_json}
\end{verbatim}
\end{tcolorbox}
\caption{Prompt template for scene-level script review.}
\label{fig:script_review_prompt}
\end{figure}

\cref{fig:script_review_prompt} shows the clip-level script review prompt used before visual generation. 
It separately evaluates the first-clip hook, last-clip ending hook, and middle-clip twist density, enabling targeted rewriting without changing unrelated parts of the scene.
\subsection{Image Review}

\begin{figure}[!h]
\centering
\begin{tcolorbox}[
    width=0.98\linewidth,
    colback=gray!5,
    colframe=blue!45!black,
    colbacktitle=blue!18,
    coltitle=black,
    title={3D-Consistent First-Frame Candidate Selection Prompt},
    fonttitle=\bfseries\centering,
    arc=2mm,
    boxrule=0.8pt,
    left=2mm,
    right=2mm,
    top=2mm,
    bottom=2mm
]
\small
You are selecting the best first frame for the next video clip.

\vspace{1mm}
You will receive exactly three images: \\
\textbf{Image 1} is the previous tail frame with the person. \\
\textbf{Image 2} is \texttt{candidate\_rank\{rank\}}'s geometry-only environment render reference. \\
Use Image 2 only for coarse spatial layout and large object relationships: walls, desks, cabinets, glass, doors, floor, and whether the person appears in a plausible spatial position. \\
Ignore Image 2's blur, noise, color, texture, material quality, lighting, and small details. \\
Do not evaluate or compare overall camera direction, camera angle, or view direction against Image 2. \\
\textbf{Image 3} is \texttt{candidate\_rank\{rank\}}, the locally refined candidate first frame to evaluate.

\vspace{1mm}
Evaluate only Image 3. Do not compare it with other candidates. \\
Use Image 1 to judge temporal continuity and character consistency. \\
Use Image 2 only to judge coarse background layout and large object spatial consistency. \\
Do not use Image 2 for color judgment. Judge color continuity only by comparing Image 3 against Image 1.

\vspace{1mm}
\textbf{Scoring rules:} \\
Score each criterion from $0$ to $5$ using integers only. \\
There are six criteria, and the maximum total score is $30$. \\
Any score below $3$ rejects the candidate, even if the total score is high. \\
Be strict, not generous. Do not give $5$ unless there is nearly no visible problem. Do not give $4$ if the flaw is clearly visible. \\
For every criterion, provide a concrete explanation in \texttt{score\_explanations}. Do not leave any explanation empty.
\end{tcolorbox}
\caption{Prompt template for 3D-consistent first-frame candidate selection.}
\label{fig:3d_candidate_selection_prompt}
\end{figure}

\begin{figure}[!h]
\centering
\begin{tcolorbox}[
    width=0.98\linewidth,
    colback=gray!5,
    colframe=blue!45!black,
    colbacktitle=blue!18,
    coltitle=black,
    title={3D-Consistent First-Frame Candidate Selection Criteria},
    fonttitle=\bfseries\centering,
    arc=2mm,
    boxrule=0.8pt,
    left=2mm,
    right=2mm,
    top=2mm,
    bottom=2mm
]
\small
\textbf{General scoring scale:} \\
$5$ = almost no issue; the criterion is essentially flawless. \\
$4$ = acceptable with only tiny flaws; usable without obvious disruption. \\
$3$ = noticeable flaw but barely usable; keep it only if other candidates are worse. \\
$2$ = unacceptable and must reject. \\
$1$ = severe failure and must reject. \\
$0$ = completely wrong, missing, or impossible to judge; must reject.

\vspace{1mm}
\textbf{Criteria:} \\
\texttt{temporal\_continuity}: Judge whether Image 3 continues naturally from Image 1 in character status, pose tendency, spatial relation, and scene continuity. \\
\texttt{layout\_consistency}: Judge whether Image 3 preserves the coarse room layout and large object relationships indicated by Image 2. \\
\texttt{background\_quality}: Judge whether the background in Image 3 is visually usable, without severe blur, broken geometry, missing regions, or distorted large objects. \\
\texttt{person\_scene\_interaction}: Judge whether the person is physically well integrated with the scene, without penetration, floating, clipping, or impossible occlusion. \\
\texttt{character\_integrity}: Judge whether the visible person in Image 3 has complete and plausible body structure, clothing, and silhouette, without severe identity or body artifacts. \\
\texttt{color\_continuity}: Judge whether Image 3 is visually compatible with Image 1 in brightness, contrast, color temperature, and overall tone.

\vspace{1mm}
\textbf{Detailed rubric for \texttt{person\_scene\_interaction}:} \\
Inspect person-scene interaction part by part before scoring: hands, fingers, feet, shoes, legs, knees, back, hips, clothing edges, hair, and the body silhouette. \\
For each body part, check whether it penetrates desks, drawers, chairs, cabinets, walls, glass panels, floor, monitors, shelves, or other objects. \\
Also check impossible occlusion order, floating limbs, clipped shoes, hands merging into drawers, back or hips sinking into furniture, and dark or transparent edge artifacts around the person. \\
$5$ = all contacts and occlusions are natural; no penetration, floating, clipping, black edge, or transparent edge artifact. \\
$4$ = mostly natural, with only tiny edge or contact imperfections. \\
$3$ = noticeable but not fatal contact issue, such as slight hand-drawer merging, odd foot contact, or local black edge. \\
$2$ = visible penetration or impossible occlusion, such as hand through drawer, shoe into floor, or back/hips sinking into furniture. Reject. \\
$1$ = multiple severe penetration/contact failures. Reject. \\
$0$ = person-environment geometry is unusable. Reject.

\vspace{1mm}
\textbf{Return strict JSON only:}
\begin{verbatim}
{
  "scores": {
    "temporal_continuity": 0,
    "layout_consistency": 0,
    "background_quality": 0,
    "person_scene_interaction": 0,
    "character_integrity": 0,
    "color_continuity": 0
  },
  "total_score": 0,
  "rejected": true,
  "score_explanations": {
    "temporal_continuity": "...",
    "layout_consistency": "...",
    "background_quality": "...",
    "person_scene_interaction": "...",
    "character_integrity": "...",
    "color_continuity": "..."
  }
}
\end{verbatim}
\end{tcolorbox}
\caption{Scoring criteria and output format for 3D-consistent first-frame candidate selection.}
\label{fig:3d_candidate_selection_criteria}
\end{figure}

\cref{fig:3d_candidate_selection_prompt,fig:3d_candidate_selection_criteria} show the prompt and scoring criteria used for selecting 3D-consistent first-frame candidates. 
The reviewer evaluates temporal continuity, coarse layout consistency, background quality, character integrity, color continuity, and person-scene interaction before video generation.
\subsection{Video Review}
\cref{fig:video_review_prompt} shows the prompt template used for generated video review. 
The reviewer checks physical realism, temporal continuity, reaction plausibility, and character presence consistency, and failed clips are sent to prompt revision or regeneration.
\begin{figure}[!h]
\centering
\begin{tcolorbox}[
    width=0.98\linewidth,
    colback=gray!5,
    colframe=blue!45!black,
    colbacktitle=blue!18,
    coltitle=black,
    title={Generated Video Review Prompt Template},
    fonttitle=\bfseries\centering,
    arc=2mm,
    boxrule=0.8pt,
    left=2mm,
    right=2mm,
    top=2mm,
    bottom=2mm
]
\small
Please score the generated video according to the following four complementary dimensions.

\vspace{1mm}
\textbf{1) \texttt{physics\_integrity}: physical realism.} \\
Check whether the video contains floating objects, gravity violations, object penetration, abnormal human body structure, duplicate entities, or other physically implausible artifacts.

\vspace{1mm}
\textbf{2) \texttt{temporal\_continuity}: temporal continuity.} \\
Check whether objects or characters flicker, disappear, teleport, abruptly change state, or miss necessary transition actions. 
Also check whether an object has already been handed over or moved but later reappears at its original position without explanation.

\vspace{1mm}
\textbf{3) \texttt{reaction\_plausibility}: reaction plausibility.} \\
Check whether character facial expressions and body actions match the scene stimulus, emotional context, and narrative situation.

\vspace{1mm}
\textbf{4) \texttt{character\_presence\_consistency}: character presence consistency.} \\
Only check whether the entrance, exit, and on-screen presence of characters match the script requirements. 
This score must be either $0$ or $10$: assign $0$ if any inconsistency exists, and assign $10$ only if it is fully consistent.

\vspace{1mm}
\textbf{General scoring scale for the first three dimensions:} \\
$0$--$2$ = severe errors; clearly unusable, with major physical violations or severe temporal jumps. \\
$3$--$4$ = many visible issues; the video feels clearly inconsistent and requires rewriting key segments. \\
$5$--$6$ = basically watchable but with obvious flaws; targeted repair is needed. \\
$7$--$8$ = mostly stable, with only minor visible issues. \\
$9$--$10$ = stable and natural, with almost no visible logical or physical problems.

\vspace{1mm}
\textbf{Additional rules for \texttt{character\_presence\_consistency}:} \\
- Characters visible at the beginning of the video should match \texttt{characters\_at\_start}. \\
- If \texttt{character\_events} is not empty, the video must show these character-set changes in the specified order. \\
- Characters visible at the end of the video should match \texttt{characters\_at\_end}. \\
- If the script says a character exits but the character remains continuously visible, \texttt{character\_presence\_consistency} must be $0$. \\
- If a character suddenly appears without a scripted entrance or narrative support, \texttt{character\_presence\_consistency} must be $0$.

\vspace{1mm}
\textbf{Return strict JSON only:}
\begin{verbatim}
{
  "physics_integrity": 0,
  "temporal_continuity": 0,
  "reaction_plausibility": 0,
  "character_presence_consistency": 0,
  "analysis": "brief evidence-based explanation"
}
\end{verbatim}
\end{tcolorbox}
\caption{Prompt template for generated video review.}
\label{fig:video_review_prompt}
\end{figure}

\clearpage
\section{Structures and Examples of Narrative Memory Units}
\label[appendix]{app:narrative_memory_units}

This appendix provides the exact structures and representative examples of
the narrative memory units introduced in
\cref{sec:Hierarchical Episode Plan}.
Our retrieval module contains two complementary offline memories.
The \emph{Pattern Bank} stores structured BeatCards that describe reusable
short-drama pacing and staging patterns, whereas the \emph{Logic Bank} stores
overlapping Logic Chunks that preserve local causal context.
After retrieval, the selected records are further abstracted into Pattern Atoms and Logic Atoms before being passed to the story writer.

The examples below are translated, lightly condensed, and anonymized to illustrate the data structures without reproducing source-specific dialogue or character names.

\subsection{Pattern Bank: From BeatCard to Pattern Atom}

Each normalized script is decomposed into an ordered sequence of BeatCards.
A BeatCard contains the fields
\texttt{beat\_id},
\texttt{beat\_type},
\texttt{hook\_type},
\texttt{opening\_action},
\texttt{closing\_hook\_visual},
\texttt{beat\_summary},
\texttt{characters},
\texttt{location}, and
\texttt{keywords}.
The cards are embedded and retrieved according to the structural needs of
the seed story:

\[
\text{script}
\rightarrow
\text{BeatCards}
\rightarrow
\text{retrieval}
\rightarrow
\text{Pattern Atom}.
\]

\begin{tcolorbox}[
    width=0.98\linewidth,
    colback=gray!5,
    colframe=blue!45!black,
    colbacktitle=blue!18,
    coltitle=black,
    title={Representative BeatCard and Its Pattern Atom},
    fonttitle=\bfseries\centering,
    arc=2mm,
    boxrule=0.8pt,
    left=2mm,
    right=2mm,
    top=2mm,
    bottom=2mm,
    breakable
]
\small

\textbf{Source category:} Wedding reversal

\vspace{1mm}
\textbf{\texttt{beat\_id}:} 1

\textbf{\texttt{beat\_type}:} Opening conflict

\textbf{\texttt{hook\_type}:} Threat

\textbf{\texttt{opening\_action}:}
The groom arrives to collect the bride and finds forty cups of liquor
arranged at the entrance.

\textbf{\texttt{closing\_hook\_visual}:}
The bride's mother blocks the doorway with both arms and announces that
no one may pass until every cup is emptied.

\textbf{\texttt{beat\_summary}:}
A ceremonial entrance is turned into a public coercion test, escalating
an interpersonal obstruction into social humiliation.

\textbf{\texttt{characters}:}
[\texttt{groom}, \texttt{bride}, \texttt{bride's mother},
\texttt{onlookers}]

\textbf{\texttt{location}:}
Villa entrance

\textbf{\texttt{keywords}:}
[\texttt{bride collection}, \texttt{forced drinking},
\texttt{public humiliation}]

\tcblower

\textbf{Retrieved abstraction---Pattern Atom}

Stage the obstruction as a physical blockade at a threshold that the
protagonist must cross, and escalate the demand in public so that refusal
carries a social cost.

\end{tcolorbox}

The Pattern Atom removes source-specific characters, dialogue, and setting
details while preserving the transferable dramatic mechanism: a physical
threshold externalizes the obstruction, and public escalation increases the
cost of refusal.

\subsection{Logic Bank: From Logic Chunk to Logic Atom}

Each normalized script is segmented into source-text windows of approximately
$900$ characters with an overlap of $120$ characters.
Each Logic Chunk stores
\texttt{doc\_id},
\texttt{source\_name}, and
\texttt{text}.
Unlike BeatCards, these chunks are not organized around predefined dramatic
beats. They retain local narrative context for resolving specific causal
questions concerning motivation, evidence activation, consequences,
relationship changes, timing, and closure:

\[
\text{script}
\rightarrow
\text{overlapping Logic Chunks}
\rightarrow
\text{retrieval}
\rightarrow
\text{Logic Atom}.
\]

\begin{tcolorbox}[
    width=0.98\linewidth,
    colback=gray!5,
    colframe=blue!45!black,
    colbacktitle=blue!18,
    coltitle=black,
    title={Representative Logic Chunk and Its Logic Atom},
    fonttitle=\bfseries\centering,
    arc=2mm,
    boxrule=0.8pt,
    left=2mm,
    right=2mm,
    top=2mm,
    bottom=2mm,
    breakable
]
\small

\textbf{\texttt{doc\_id}:}
\texttt{[anonymized]}

\textbf{\texttt{source\_name}:}
\texttt{[anonymized source script]}

\textbf{\texttt{text} (translated and condensed):}
A consort lies motionless after making a life-saving sacrifice.
Only at the moment of her death does the beneficiary learn that she once
saved him, leaving no opportunity for gratitude or repayment.

\vspace{1mm}
\textbf{Retrieval query:}
What makes a past sacrifice a credible motive for revenge?

\tcblower

\textbf{Retrieved abstraction---Logic Atom}

A past sacrifice becomes a credible motive for revenge when it is revealed
too late to repay: the beneficiary learns of it at the moment of loss, so
the debt can never be settled.

\end{tcolorbox}

The Logic Atom converts the retrieved local context into a causal constraint
for story drafting. In this example, the motivation does not arise from the
sacrifice alone; it depends on the delayed revelation, the irreversible loss,
and the impossibility of repayment. Thus, BeatCards and Pattern Atoms provide
reusable pacing and staging priors, while Logic Chunks and Logic Atoms provide
local causal support. Only the abstracted atoms, rather than the retrieved
source records, are passed to the downstream story-generation stage.

\section{Personalization from User-Provided Portraits}
\label[appendix]{app:personalization}

Our framework is personalized at two levels: the drama is generated from the user's own single-sentence idea, and the user may cast themselves or anyone they choose as the protagonists by uploading a portrait.

Concretely, the scene-level character assets in \cref{sec:visual assets and prompt generation} accept either a generated seed portrait or a user-uploaded photo. From an uploaded portrait, we generate multi-view character references under the wardrobe descriptions from the story core, and these references are then used throughout the pipeline: conditioning first-frame synthesis, providing the appearance transferred onto the rendered mesh in the 3D-grounded branch, and serving as the character-reference elements passed to the video generator. No fine-tuning or per-user training is required, and uploaded and generated identities are treated identically, so a user can supply portraits for only some characters and let the framework generate the rest.

\end{document}